\documentclass[pdflatex,sn-mathphys-num,iicol]{sn-jnl}

\usepackage{graphicx}%
\usepackage{multirow}%
\usepackage{amsmath,amssymb,amsfonts}%
\usepackage{amsthm}%
\usepackage{mathrsfs}%
\usepackage[title]{appendix}%
\usepackage{xcolor}%
\usepackage{textcomp}%
\usepackage{manyfoot}%
\usepackage{booktabs}%
\usepackage{algorithm}%
\usepackage{algorithmicx}%
\usepackage{algpseudocode}%
\usepackage{listings}%
\usepackage{graphicx}
\usepackage{multirow}
\usepackage{pifont}
\usepackage{booktabs}
\usepackage{CJKutf8}
\usepackage{color}
\usepackage{makecell}
\usepackage{subfig}
\usepackage{graphicx}

\theoremstyle{thmstyleone}%
\theoremstyle{thmstyletwo}%

\theoremstyle{thmstylethree}%

\begin{document}

\title[Article Title]{IPAD: Iterative, Parallel, and Diffusion-based Network for Scene Text Recognition}


\author[1,2]{\fnm{Xiaomeng} \sur{Yang}}\email{yang.xiaome@northeastern.edu}

\author[3]{\fnm{Zhi} \sur{Qiao}}\email{qiaozhi1@tal.com}

\author*[1]{\fnm{Yu} \sur{Zhou}}\email{yzhou@nankai.edu.cn}

\affil*[1]{\orgdiv{VCIP \& TMCC \& DISSec, College of Computer Science}, \orgname{Nankai University}, \orgaddress{\city{Tianjin}, \postcode{300350}, \country{China}}}

\affil[2]{\orgdiv{Institute of Information Engineering}, \orgname{CAS}, \orgaddress{\city{Beijing}, \postcode{100089}, \country{China}}}

\affil[3]{\orgdiv{Tomorrow Advancing Life}, \orgaddress{\city{Beijing}, \postcode{100080}, \country{China}}}


\abstract{Nowadays, scene text recognition has attracted more and more attention due to its diverse applications. Most state-of-the-art methods adopt an encoder-decoder framework with the attention mechanism, autoregressively generating text from left to right. Despite the convincing performance, this sequential decoding strategy constrains the inference speed. Conversely, non-autoregressive models provide faster, simultaneous predictions but often sacrifice accuracy. Although utilizing an explicit language model can improve performance, it burdens the computational load. Besides, separating linguistic knowledge from vision information may harm the final prediction. In this paper, we propose an alternative solution that uses a parallel and iterative decoder that adopts an easy-first decoding strategy. Furthermore, we regard text recognition as an image-based conditional text generation task and utilize the discrete diffusion strategy, ensuring exhaustive exploration of bidirectional contextual information. Extensive experiments demonstrate that the proposed approach achieves superior results on the benchmark datasets, including both Chinese and English text images.}




\maketitle

\section{Introduction}\label{sec1}

Nowadays, the field of scene text detection~\cite{zhou2017east} and recognition~\cite{shi2016end} is drawing substantial research attention because of its significant role in many downstream tasks, such as text spotting~\cite{wang2022tpsnet} and text-based visual question answering~\cite{zeng2023beyond}. Scene text detection involves the localization of text instances in scene images, and scene text recognition (STR) transcribes the localized instances into an editable text format. STR, in particular, is laden with challenges stemming from the diversity of characters and backgrounds. However, the advent and progression of deep learning have facilitated the achievement of convincing results in this domain. 

Based on the decoding strategy, existing STR methods can be roughly categorized into three distinct types: Connectionist Temporal Classification (CTC) based~\cite{he2016reading,shi2016end,su2017accurate,wang2017gated,du2022svtr}, attention mechanism based~\cite{shi2018aster,baek2019wrong,li2019show,yu2020towards,bautista2022scene} and segmentation-based~\cite{liao2019scene,wan2020textscanner} methods. From another perspective, these methods can also be classified into autoregressive and non-autoregressive approaches. Specifically, autoregressive methods decode the text from left to right, and the number of regressions depends on the length of the text. Most attention mechanism-based methods adopt a left-to-right autoregressive decoding process. Recently, autoregressive methods have achieved great success in scene text recognition~\cite{li2019show,zhong2022sgbanet,bautista2022scene,yang2024masked}.

Although the performance is satisfactory, the inference speed of autoregressive models is relatively slow, especially when dealing with long texts. On the contrary, non-autoregressive methods predict the text in parallel, with the CTC-based methods serving as typical exemplars. Such parallel inference improves the speed significantly but ignores the dependencies between characters. We argue that fully non-autoregressive methods lack the context information of characters, which is significant to the recognition of hard cases, and the assumption of independence increases the training difficulties of the hidden layers. 
Compared with autoregressive methods, the accuracy of non-autoregressive methods is relatively poor. An alternative approach involves supplementing non-autoregressive models with an explicit language module, refining predictions independently from visual data~\cite{yu2020towards, fang2021read}. However, this approach divides vision and language processing, which leads to a lack of visual information within the language model, so it can only adjust the results according to the pretrained language weight from the training text and may adjust the correct prediction from the vision model to a wrong one. Besides, this external language model-based solution poses substantial computational demands.

In this paper, we propose a parallel and iterative decoding model. In each iteration, the decoder still generates the text in parallel, and the context information is extracted depending on the previous predictions. Specifically, we adopt the easy-first~\cite{goldberg2010efficient} decoding strategy according to the iterative generative process. Easy-first strategy predicts the most confident and obvious characters in each iteration first and re-predicts other remaining characters in the next iteration. Unlike traditional left-to-right decoding, easy-first breaks the limitation of decoding order with higher flexibility. Inspired by Transformer~\cite{vaswani2017attention}, the parallel decoder consists of a masked self-attention module, a 2D cross-attention module, and a feed-forward network (FFN).

Furthermore, although the iterative prediction brings the context information into the decoding, the pre-defined easy-first decoding training strategy can only allow the parallel decoder to exploit limited context according to the decoding order of the training dataset. Since the quantity of the training dataset is limited, the inter-character dependence may not be comprehensively contained in this scope. Therefore, it is more suitable during inference instead of training.
To solve this problem, we contend that STR can be interpreted as a conditional text generation task depending on the input image. Therefore, we can employ the procedure of the discrete diffusion model for the STR task. Diffusion models~\cite{dhariwal2021diffusion,rombach2022high, he2022diffusionbert} have demonstrated exceptional capabilities in tasks involving image and text generation, which is a potential solution. Their superior semantic learning capabilities and ability to integrate information across multiple stages of the decoding process could effectively address the complexities of STR. To achieve this, we adopt the discrete diffusion proposal into our parallel and iterative decoding framework. 
During training, the forward noising process is applied with a randomly chosen timestep, and the clean text is noised by replacing the characters with placeholders. In the denoising process, our decoder directly predicts the clean text. For inference, the denoising process is combined with our easy-first decoding process, which means the re-noising process is determined by the predicted confidence instead of the predefined noising process. In this paper, we call our proposed method an Iterative, Parallel, and Diffusion-based Network (IPAD), and the main framework of IPAD is shown in Fig.~\ref{fig_architecture}.

To be specific, in this paper, we propose an advanced framework that is different from our conference version PIMNet~\cite{qiao2021pimnet}\footnote{A best paper candidate in ACM MM 2021.} in the following aspects: 
(1) We deeply analyze the reason behind the performance gap between non-autoregressive and autoregressive models and propose to view the STR task as an image-based text-generation task and adopt a discrete diffusion training strategy for the STR task. 
With the training of the denoising process for numerous noise texts used at different timesteps, our parallel and iterative decoder could learn sophisticated linguistic knowledge implicitly. 
(2) Different from English, the context of Chinese is more complex and the length of the Chinese text is longer, thus recognizing the Chinese text needs more advanced contextual information. Our method can learn the relationship between characters through easy-first decoding and diffusion training, which is more effective in Chinese text recognition. We implement more experiments and analyses, demonstrating that the proposed methods are effective and superior both on Chinese and English datasets.

The main contributions of our work are as follows:
\begin{itemize}
  \item 
We propose a parallel and iterative decoding framework for text recognition. Different from previous works, our method decodes with constant iterations independent of the text length, which takes advantage of both fully autoregressive and fully non-autoregressive methods and achieves a good balance between accuracy and efficiency.
  
  \item 
To achieve parallel and iterative decoding, an easy-first strategy is designed for text recognition. Different from traditional
left-to-right decoding, the easy-first strategy predicts the most confident characters in each iteration, which is more flexible. 

\item 
To enhance the ability of contextual learning and help the training of the parallel decoder, we employ the discrete diffusion training strategy, which views the text recognition process as an image-based conditional text generation. 

\item 
Extensive experiments are conducted to verify the effectiveness and efficiency of the proposed method, which achieves state-of-the-art or comparable accuracies on six popular benchmarks and three large datasets with a faster inference speed. Besides, it also achieves state-of-the-art performance on the Chinese benchmark datasets.

\end{itemize}

\section{Related Work}
\subsection{Scene Text Recognition}
Scene text recognition has been studied for many years, and existing methods can be classified into two primary categories: traditional methods and deep learning-based methods. Traditional methods~\cite{mishra2012scene,mishra2012top,neumann2012real,novikova2012large,wang2011end,wang2010word,wang2012end,yao2014strokelets} usually employ a bottom-up framework, which initially detects and classifies characters, then subsequently group them utilizing a lexicon, a language model or heuristic rules. In recent years, deep learning-based methods have dominated this area because of the simple pipeline and compelling performance. Moreover, these methods can be further bifurcated, depending on their prediction strategies, into two groups: non-autoregressive and autoregressive methods as follows.

\subsection{Autoregressive Text Recognition}
Autoregressive methods typically adopt an encoder-decoder framework that predicts sequences from left to right. Most attention-based methods belong to the autoregressive model and can be divided into 1D attention-based and 2D attention-based. For 1D attention-based, Lee and Osindero~\cite{lee2016recursive} proposes a recursive CNN network to capture broader features and an attention-based decoder to transcribe sequence. Cheng et al.~\cite{cheng2017focusing} introduces the problem of attention drift and proposes a focusing attention network to solve it. Fang et al.~\cite{fang2018attention} proposes a fully CNN-based network to extract visual and language features separately. Some methods~\cite{luo2019moran,shi2016robust,shi2018aster} rectify irregular text image first then recognize it with 1D attention-based decoder. ESIR~\cite{zhan2019esir} and ScRN~\cite{yang2019symmetry} improve the quality of rectification with iteration and additional geometrical constraints, respectively. 
For 2D attention-based, Yang et al.~\cite{yang2017learning} first introduces 2D attention into irregular text recognition, proposing an auxiliary segmentation task. Li et al.~\cite{li2019show} suggests a tailored 2D attention operation.

Despite their robust performance resulting from the integration of preceding contextual information into the decoding process, autoregressive methods do face a set of challenges. Specifically, they encounter the issue of attention drift for long texts, and their one-way serial decoding process restricts the amount of contextual information that can be considered.
Addressing the challenge of attention drift, DAN~\cite{wang2020decoupled} offers a solution that decouples the prediction of attention weights from the decoding process, and Qiao et al.~\cite{qiao2021gaussian} introduces a Gaussian constrained refinement module to refine the attention distribution. Further developments focus on employing advanced linguistic knowledge to improve the decoding process. SEED~\cite{qiao2020seed} introduces semantic global information to guide the decoding process. SCATTER~\cite{litman2020scatter} trains a deep BiLSTM encoder to extract broader contextual dependencies. Zheng et al.~\cite{zheng2020lal} adopts an external language model to incorporate useful context information into the decoding. PARSeq~\cite{bautista2022scene} learns an internal language model through permutations of autoregressive decoding. PTIE~\cite{tan2022pure} trains the ViT encoder and transformer decoder with varying patch resolutions and AR decoding directions.

\subsection{Non-Autoregressive Text Recognition}
Non-autoregressive methods aim to generate the target text in a single iteration or a constant time independent of the text length. Jaderberg et al.~\cite{jaderberg2016reading} interprets word recognition as a classification task using CNN, which suffers from scalability issues due to a fixed vocabulary. ViTSTR~\cite{atienza2021vision} adapts the Vision Transformer (ViT)~\cite{dosovitskiy2020image} for scene text recognition, which directly classifies each visual representation learned by the ViT encoder in parallel. Many methods treat text recognition as a sequence-to-sequence task, and non-autoregressive decoding generally relies on one of the three major technologies: CTC-based, segmentation-based, and parallel attention-based. A majority of the CTC-based methods~\cite{chao2020variational,feng2019textdragon,he2016reading,hu2020gtc,shi2016end,su2017accurate,wang2017gated} utilize a CNN to extract visual features and CTC to transcribe the final text with a short inference time. SVTR~\cite{du2022svtr} further enhances this by designing local and global mixing blocks for a ViT-based encoder, capturing more refined visual features for the CTC-based decoder. Segmentation-based methods~\cite{liao2019scene,wan2020textscanner,guan2022glyph} regard text recognition as a task of semantic segmentation of characters, and they need additional character-level annotations. 

Parallel attention-based methods have gained widespread attention in various fields, such as neural machine translation~\cite{ghazvininejad2019mask,gu2017non,qian-etal-2021-glancing}, automatic speech recognition~\cite{chan2020imputer,tian2020spike,chi-etal-2021-align} and image caption~\cite{guo2020non}, thanks to their inference speed. Many recent non-autoregressive STR methods are grounded in this parallel attention decoder framework. 
However, one significant limitation of the complete one-time parallel decoding methods is that they typically assume each character to be independent. This assumption hinders their ability to utilize contextual information during the decoding process, which is especially problematic when recognizing difficult cases. Therefore, parallel attention-based scene text recognition methods have introduced various strategies to incorporate linguistic knowledge into their models.

For instance, SRN~\cite{yu2020towards} and ABINet~\cite{fang2021read} design an independent semantic module following parallel visual attention. This arrangement allows the semantic module to learn contextual information and refine initial predictions made by the vision module. LevOCR~\cite{da2022levenshtein} further explores the effective fusion of visual and linguistic features. However, the decoupled two-stage model structure makes context information limited by the predictions of the parallel visual attention module, and the context information may tend to accumulate errors due to the wrong predictions. To circumvent this, VisionLAN~\cite{wang2021two} introduces a visual reasoning module that randomly masks corresponding visual features of characters during training, enabling the implicit learning of linguistic knowledge within vision space. Furthermore, MGP-STR~\cite{wang2022multi} includes subword representations to enable multi-granularity prediction. 

In summary, while autoregressive models typically produce superior accuracy, their sequential generation approach constrains inference speed. Conversely, non-autoregressive models offer enhanced inference speed, but often at the expense of lower accuracy. Even though incorporating linguistic knowledge improves the performance of non-autoregressive methods, these usually require an additional semantic refinement module or a multi-stage training procedure. In contrast, our work seeks a balance between fully autoregressive and fully non-autoregressive models. Our objective is to explore the development of a straightforward yet powerful iterative decoding model that achieves an optimal balance between efficiency and accuracy. Furthermore, we intend to equip our model with internal linguistic knowledge to boost its performance further.

\subsection{Diffusion Models}
Diffusion models, characterized by a forward noising process and a reverse denoising operation, have recently been classified as the current state-of-the-art in the family of deep generative models~\cite{croitoru2023diffusion,yang2022diffusion}. They function by incrementally injecting noise into clean samples in the forward process and then employing the reverse process to counteract this corruption, recovering the original samples. These models can be categorized into continuous diffusion models and discrete diffusion models.

Continuous diffusion models~\cite{dhariwal2021diffusion,ho2020denoising,song2020denoising,zhang2022gddim}, which introduce noise into continuous-valued input or latent features, have been extensively applied in various tasks involving generation. They have achieved remarkable results in image generation~\cite{nichol2021glide,ramesh2022hierarchical,saharia2022photorealistic,rombach2022high}, video generation~\cite{ho2022video,ho2022imagen}, audio generation~\cite{kong2020diffwave} and 3D generation~\cite{xu2022dream3d}. Despite the inherent discreteness of text, researchers have made efforts to employ continuous diffusion models for text generation. These methods~\cite{li2022diffusion,gong2022diffuseq} utilize designed mapping functions to bridge the gap between the continuous and discrete domains. 

Regarding models of discrete diffusion~\cite{sohl2015deep}, each element is treated as a discrete random variable with several categories, and the diffusion process happens within discrete states. ImageBART~\cite{esser2021imagebart} introduces an innovative application of multinomial diffusion on the latent discrete code space of a VQ-VAE~\cite{Razavi2019GeneratingDH}, learning a parametric model for image generation. VQ-Diffusion~\cite{gu2022vector} alternatively substitutes Gaussian noise with a random walk in the discrete data space. For text generation, Hoogeboom et al.~\cite{hoogeboom2021argmax} puts forward the use of multinomial diffusion for character-level text generation, employing the forward categorical noise through the Markov transition matrix. D3PM~\cite{austin2021structured} broadens the scope of discrete text diffusion models by introducing the $\langle {\rm MASK}\rangle$ token. Different from preceding methods, DiffusionBERT~\cite{he2022diffusionbert} integrates pretrained language models with absorbing-state discrete diffusion models for text. Additionally, within multi-modal tasks, DDCap~\cite{zhu2022exploring} deploys the vector quantized discrete diffusion models for image captioning.

We propose viewing the text recognition task as a form of conditional text generation. In this paper, given the discrete nature of characters in texts, we employ the capabilities of the discrete diffusion model in scene text recognition. This approach trains a stochastic process and empowers the decoder to learn a suite of distributions. Consequently, it can progressively recognize and refine results, informed by the internal linguistic knowledge acquired during training.

\begin{figure*}[!t]
\centering
\subfloat[PIMNet~\cite{qiao2021pimnet} Architecture]{\includegraphics[width=4in]{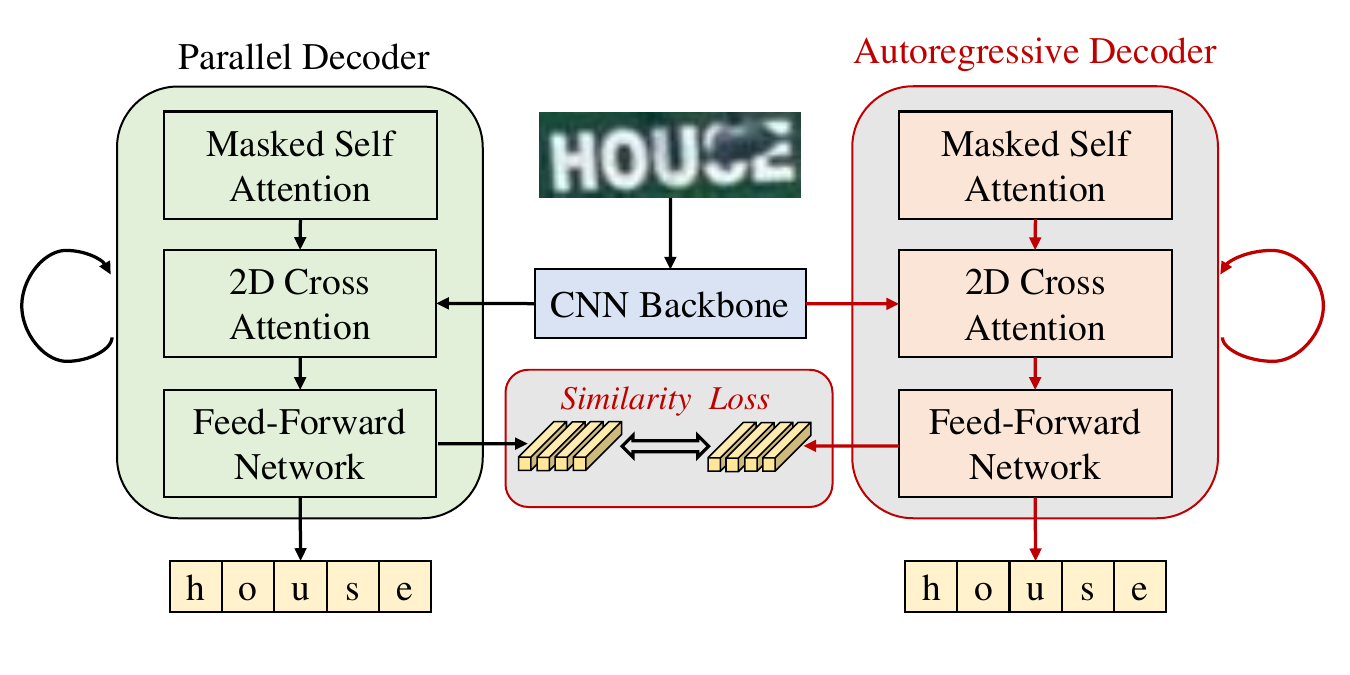}%
\label{pimnet_architecture}}
\hfil
\subfloat[IPAD Architecture]{\includegraphics[width=4in]{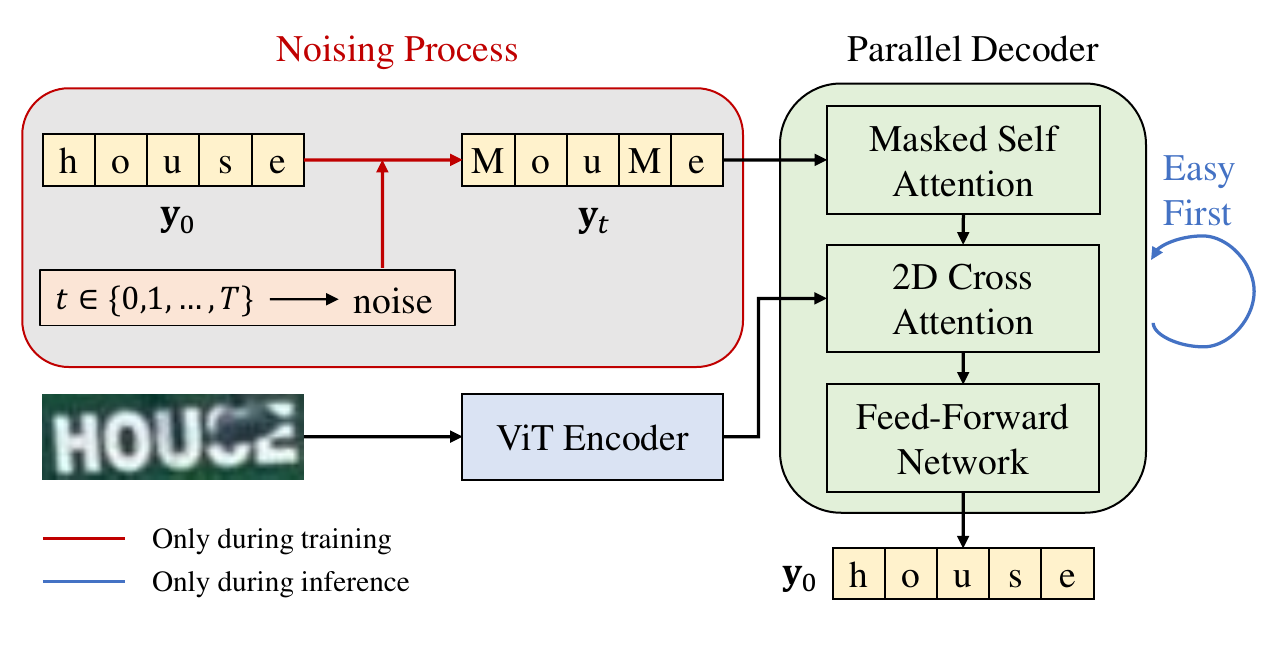}%
\label{ipad_architecture}}
\caption{Comparison between the proposed IPAD and our conference version PIMNet~\cite{qiao2021pimnet}. For IPAD, the ViT-based encoder first extracts the visual feature, and then the parallel decoder adopts an iterative generation to extract the context information from the previous predictions. The discrete diffusion strategy is used during training, where the noising process adds noise by randomly sampling a timestep and then the parallel decoder works as the denoising network, which directly predicts the clean text without any iteration.}
\label{fig_architecture}
\end{figure*}

\section{Methodology}
In this section, we will describe our proposed model in detail. As depicted in Figure~\ref{fig_architecture}, our model consists of two principal components: a ViT-based Encoder and a parallel decoder. To break the assumption of character independence in the fully parallel decoder, the easy-first decoding strategy is employed to predict the text iteratively during inference. Besides, to facilitate the internal contextual information learning of our parallel decoder, we incorporate a discrete diffusion procedure and make the parallel decoder denoise the text to clean text directly during training.
In Section~\ref{encoder}, we first introduce the encoder employed for visual feature extraction. Subsequent sections delineate the iterative decoding process and parallel decoder in Section~\ref{Decoding} and Section~\ref{Decoder}, respectively. Section~\ref{Diffusion} describes the concept of deploying discrete diffusion models for STR decoding. Finally, in Section~\ref{Training}, we will outline the training and inference process, as well as the objective function.

\subsection{ViT-based Encoder} \label{encoder}
Contrasting with our conference version, which employs a ResNet50 equipped with FPN~\cite{he2016deep} as the encoder, we opt for a ViT-based~\cite{dosovitskiy2020image} encoder in this work, considering its capability to provide sophisticated visual features. Same as PARSeq~\cite{bautista2022scene} and MGP-STR~\cite{wang2022multi}, we select the 12-layer ViT as our encoder. The classification head and the $\langle {\rm CLS}\rangle$ token are ignored. When an image $\mathbf{x}\in \mathbb{R}^{\mathcal{H}\times \mathcal{W}\times \mathcal{C}}$ is input, the encoder partitions it into $\mathcal{N} = \mathcal{H}\mathcal{W}/(\mathcal{P}_w\mathcal{P}_h)$ patches of shape $\mathcal{P}_w\times \mathcal{P}_h$. These patches are then linearly projected into $\mathcal{D}$-dimensional tokens, together with the learned position embeddings, thus generating a sequence of visual features $\mathbf{z}\in \mathbb{R}^{\mathcal{N}\times \mathcal{D}}$ for the decoder.

\begin{figure}[!t]
\centering
\includegraphics[width=0.95\columnwidth]{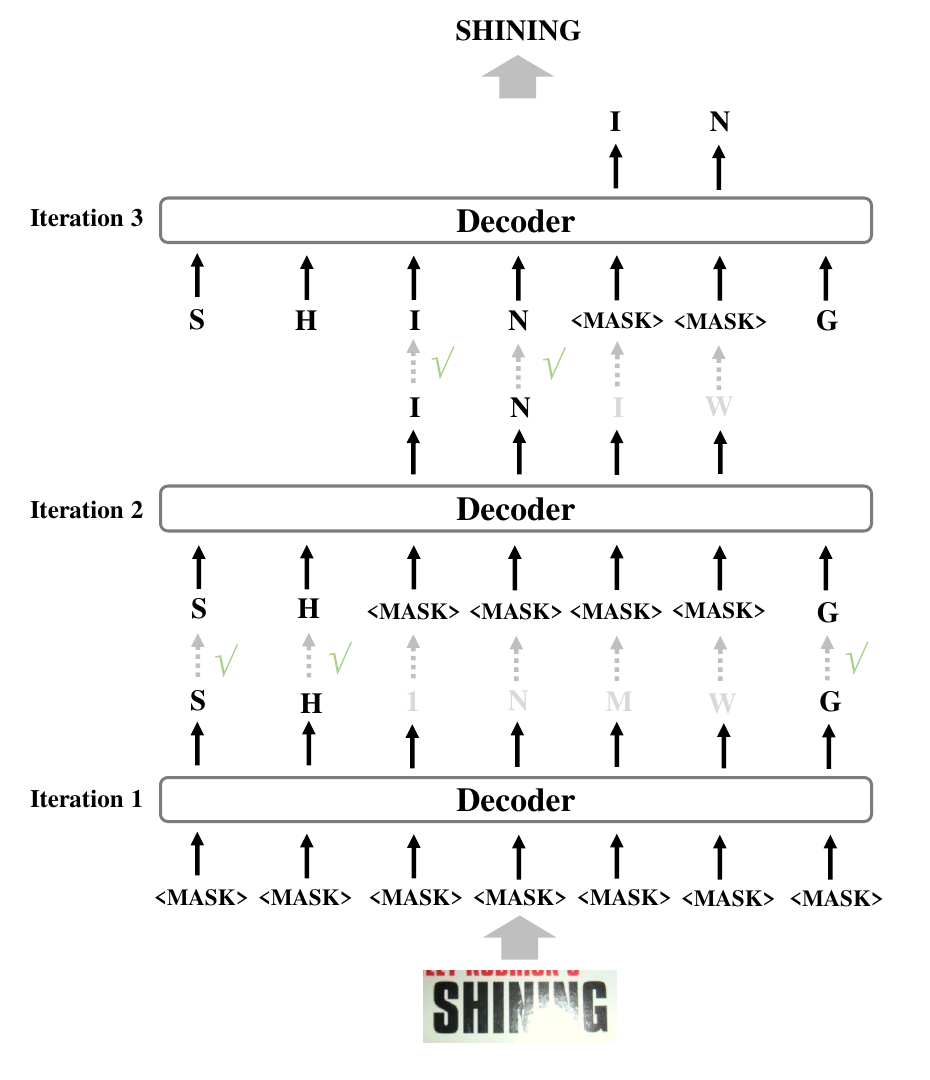}
\caption{An illustration of the easy-first decoding strategy. In each iteration, the characters in black represent the characters with high confidence and will be reserved in the next iteration. The characters with low confidence will be replaced with the $\langle {\rm MASK}\rangle$ token and re-predicted again based on the other reachable predictions}
\label{fig_easy_first}
\end{figure}

\subsection{Iterative Decoding Strategy} \label{Decoding}

Easy first~\cite{goldberg2010efficient} is an iterative decoding strategy where the most confident predictions in each iteration are first predicted. Inspired by \cite{devlin2018bert}, we adopt a particular token $\langle {\rm MASK}\rangle$ for the iterative decoding, which acts as a placeholder for the next iteration. As shown in Figure~\ref{fig_easy_first}, all characters are unreachable in the first iteration, so the input to all of the decoder's positions is $\langle {\rm MASK}\rangle$ token. The next iterations can be separated into two main steps of prediction and update:

\textbf{Prediction}: In the current iteration, the parallel decoder predicts the corresponding probabilities of characters for each position which is still $\langle {\rm MASK}\rangle$. On the contrary, the characters which have been already updated will not be re-predicted again:
\begin{equation}
{\hat{y}^l_i} = \begin{cases}
\mathop{\rm argmax}\limits_{c_n\in C}(P(c^l_n|y_{i-1})),&{\text{if}}\ {y_{i-1}^l \text{is}\ \langle {\rm MASK}\rangle}; \\ 
{y_{i-1}^l},&{\text{otherwise}}. 
\end{cases}
\end{equation}
where $y_{i-1}$ is the character sequence in the previous iteration $i-1$, $\hat{y}^l_i$ indicates the predicated character in position $l\in [1, 2, ..., L]$ in iteration $i$, $P(c^l_n)$ is the predicted probability for the character $c_n$ in position $l$, and $argmax(P)$ chooses the character with the largest probability. Here $C = \{c_1, c_2, ..., c_{N+1}\}$ is the meaningful $N$-charset added with the $\langle {\rm EOS}\rangle$ token. The conditional probability in $P$ indicates that the context information has been fully utilized based on the predictions from the previous iteration in a bi-directional manner.

\textbf{Update}: After prediction, some positions of the target text are updated with the most confident predictions, and the others are abandoned and replaced by $\langle {\rm MASK}\rangle$ again:
\begin{equation} \label{eq:update}
{y^l_i} = \begin{cases}
\hat{y}^l_i,&{\text{if}}\ {l\in {\rm top}_k(\mathop{\rm argmax}\limits_{c_n\in C}(P(c^M_n | y_{i-1})))}; \\ 
{\langle {\rm MASK}\rangle},&{\text{otherwise}}.
\end{cases}
\end{equation}
where $y_i$ is the final prediction in this iteration and will act as the input in the next iteration, $max(P)$ is the maximum probability for all candidate characters, $M$ indicates all the $\langle {\rm MASK}\rangle$ positions in the $(i-1)$th iteration. $top_k$ represents the most confident $k$ predictions in the $M$ positions which are picked to update the previous predictions. In our setting, $k$ is equal to $L/K$ where $L$ is the max length of the target text, and $K$ is the iteration number.
    
In summary, the predicted text is updated at the most $top_k$ confident positions in each iteration, and the others are replaced by $\langle {\rm MASK}\rangle$ again, which will be predicted in the following iterations. Once a character is reserved and updated, we will not update it again in the subsequent iterations. We utilize the same post-processing method to deal with the length determination described in our conference version~\cite{qiao2021pimnet}, where we let the decoder predict $\langle {\rm EOS}\rangle$ token and choose the first $\langle {\rm EOS}\rangle$ token from left to determine the text length.

\subsection{Parallel Decoder} \label{Decoder}
We adopt a Transformer-based \cite{vaswani2017attention} decoder as our parallel decoder without any RNN structure for high parallelism. It contains three main components: a masked self-attention, a 2D cross-attention, and an FFN. The input of masked self-attention is the text embedding combined with position encoding, which is mixed with the $\langle {\rm MASK}\rangle$ tokens. Different from the original Transformer, our parallel decoder is bidirectional, thus removing the original future-characters mask for the left-to-right generation. However, we add another mask to prevent the model from attending to the special $\langle {\rm MASK}\rangle$ tokens. The masked self-attention can extract abundant context information ignored in the fully non-autoregressive models.

The 2D cross-attention is similar to the original Transformer, where the multi-head scale-dot attention operation is adopted. Specifically, the attention weights are calculated between the outputs of masked self-attention and the 2D visual feature map provided by the vision encoder. The calculation details are the same as the Transformer, so we don’t describe it in detail. Finally, the FFN introduces the non-linear transformation into the outputs of 2D cross-attention, and new predictions are generated later with a simple linear function in parallel. Same as the original Transformer, the residual connections exist between each module. 

\subsection{Discrete Diffusion for STR} \label{Diffusion}
In our conference paper~\cite{qiao2021pimnet}, we utilize the mimicking learning strategy to enhance the training of the parallel decoder, which makes the parallel decoder mimic the FFN outputs of the teacher model with an autoregressive decoder. However, the autoregressive decoder has its intrinsic defect. Its left-to-right serial language modeling strategy ignores the bidirectional contextual information and suffers from error accumulation. Therefore, we introduce another strategy, the discrete diffusion model strategy, which could learn abundant bidirectional linguistic knowledge to enhance internal language model learning. 

The easy first decoding could be viewed as a kind of iterative denoising process of a discrete diffusion model. Here, the corrupted text $y_T$ is formed by all $\langle {\rm MASK}\rangle$ tokens, and the denoising process needs to recover the correct text conditioned on the image visual information, which means that the scene text recognition task can be viewed as a conditional text-generation task, i.e., we need to generate (recognize) the text from the given image. Unlike the continuous diffusion models, which add continuous noises individually to the RGB values or latent features of each token, the discrete character token of a word determines it is direct to model the process as the vector quantized diffusion~\cite{gu2022vector}. In the forward process, our model masks the character tokens randomly by replacing the original characters with $\langle {\rm MASK}\rangle$ tokens according to the randomly chosen timestep and masking ratio. In the backward process, our model recovers the original texts from the corrupted masked texts conditioned on the input image.

\subsubsection{Noising Process}
The noising process can be thought of as the Markov process, where the forward noise process gradually corrupts the text $\mathbf{y}_0$ via a fixed Markov chain $q(\mathbf{y}_t|\mathbf{y}_{t-1})$. To be specific, the original text $\mathbf{y}_0$ is the word contained in the image, which is formed by the tokens in $C = \{c_1, c_2, ..., c_{N+1}\}$ consisting of the meaningful $N$ character categories and the special $\langle {\rm EOS}\rangle$ token. At each timestep, we randomly replace some unmasked tokens of $\mathbf{y}_{t-1}$ with the special token $\langle {\rm MASK}\rangle$ with the probability $\beta_t$, leaving $1-\beta_t$ to be unchanged. The tokens never change to other unmasked tokens, and the special $\langle {\rm MASK}\rangle$ never changes to other tokens. For a character $y^l_{t-1}$ of $\mathbf{y}_{t-1}$ in position $l$ (omit the superscripts $l$ in the following description), if it is not $\langle {\rm MASK}\rangle$, the transition is defined as:
\begin{equation}
{q(y_t|y_{t-1})} = \begin{cases}
1-\beta_t,&{\text{if}}\ {y_t = y_{t-1}}; \\ 
{\beta_t},&{\text{if}}\ {y_t = \langle {\rm MASK}\rangle};\\
0,&{\text{otherwise}}.
\end{cases}
\end{equation}
If $y_{t-1}$ is a $\langle {\rm MASK}\rangle$ token, the transition is defined as:
\begin{equation}
{q(y_t|y_{t-1})} = \begin{cases}
1,&{\text{if}}\ {y_{t-1} = \langle {\rm MASK}\rangle}; \\ 
{0},&{\text{otherwise}}.
\end{cases}
\end{equation}
We define $c_{N+2} = \langle {\rm MASK}\rangle$. Denote $[\mathbf{Q}_t]_{ij} = q(y_t=c_i|y_{t-1}=c_j)\in \mathbb{R}^{(N+2)\times (N+2)}$ as the transcation matrix at timestep t,
\begin{equation}
\mathbf{Q}_t = 
\begin{bmatrix} 
1-\beta_t & 0 & 0 & \cdots & 0\\ 
0 & 1-\beta_t & 0 & \cdots & 0\\
0 & 0 & 1-\beta_t & \cdots & 0\\
\vdots & \vdots & \vdots & \ddots & \vdots\\
\beta_t & \beta_t & \beta_t & \cdots & 1
\end{bmatrix}
\end{equation}
Then the forward Markov diffusion process for the whole token sequence can be written as:
\begin{equation}
q(\mathbf{y}_t|\mathbf{y}_{t-1}) = \mathbf{v}^{\top}(\mathbf{y}_t)\mathbf{Q}_t\mathbf{v}(\mathbf{y}_{t-1})
\end{equation}
$\mathbf{v}(\mathbf{y})$ is a one-hot column vector representation of $\mathbf{y}$ with the length of $(N+2)$. According to the property of the Markov chain rule, we can derive the corrupted text $\mathbf{y}_t$ from $\mathbf{y}_0$ as:
\begin{equation}
q(\mathbf{y}_t|\mathbf{y}_0) = \mathbf{v}^{\top}(\mathbf{y}_t)\overline{\mathbf{Q}}_t\mathbf{v}(\mathbf{y}_{t-1})
\end{equation}
where $\overline{\mathbf{Q}}_t=\mathbf{Q}_t\dots\mathbf{Q}_1$. More specially, denote the $\alpha_t = 1-\beta_t$ and $\overline{\alpha}_t = \prod_{i=1}^t\ \alpha_i$,
\begin{equation}
q(y_t|y_0) = 
\begin{cases}
\overline{\alpha}_t, &{\text{if}}\ {y_t = y_0}; \\ 
{1 - \overline{\alpha}_t}, &{\text{if}}\ {y_t = \langle {\rm MASK}\rangle}; \\
{0}, &{\text{otherwsie}}.
\end{cases}
\end{equation}
After a fixed number of $T$ timesteps, the probability of $\overline{\alpha}_t$ turns close to $0$, and the noise process yields a sequence of pure noise tokens.

\subsubsection{Conditional Denoising Process}
The recognition process is the conditional denoising process from a sequence of all masked characters $\mathbf{y}_T$ to the original text $\mathbf{y}_0$ based on the visual features $\mathbf{z}$ provided by the encoder. To reverse the diffusion process, we train a denoising network $p_\theta(\mathbf{y}_{t-1}|\mathbf{y}_t,\mathbf{z})$ to estimate the posterior transition distribution. The denoising network used in our model is the Transformer-based parallel decoder described in Section~\ref{Decoder}. The conditional image features are incorporated through the 2D cross-attention.

Since the noising and denoising process happens at the character level of texts in scene images, the timescale $T$ in our task is significantly small compared with the popular image generation task. We adopt a scaling between our $T$ and the popular used $8,000$ timestep in the image generation task, i.e., $t_s = t * 8000/T$. To indicate current timestep $t$, we encode it as a sinusoidal positional embedding and inject it to the parallel decoder with the Adaptive Layer Normalization~\cite{ba2016layer}:
\begin{equation}
\text{PE}(t_s) = 
\begin{cases}
{\rm sin}(t_s/10000^{2i/d_{model}}), &{\text{if}}\ {i < d_{model}/2}; \\ 
{\rm cos}(t_s/10000^{2i/d_{model}}), &{\text{if}}\ {i \ge d_{model}/2}.
\end{cases}
\end{equation}
\begin{equation}
\text{AdaLN}(h, t_s) = W_1\text{PE}(t_s)\text{LayerNorm}(h) + W_2\text{PE}(t_s)
\end{equation}
where $h$ is the intermediate activations, $W_1$ and $W_2$ are learnable linear projection weights of the timestep embedding.

\subsection{Training and Inference Strategies} \label{Training}
In the training process, the decoder directly predicts the original noiseless text tokens $\mathbf{y}_0$ given the generated noise text $\mathbf{y}_t$ according to the exploration of \cite{austin2021structured}. The objective function is just like the usual text recognition loss:
\begin{equation}
    \mathcal{L}_{x_0} = -{\rm log}\ p_\theta(\mathbf{y}_0|\mathbf{y}_t, \mathbf{x})
\end{equation}

The standard diffusion inference process starts from the $\mathbf{y}_T$, and sequently predict the $\mathbf{y}_{T-1}, \mathbf{y}_{T-2}, ..., \mathbf{y}_{0}$. And the process of predicting $\mathbf{y}_{t-1}$ from $\mathbf{y}_t$ is computed as follows: First, estimate $\hat{\mathbf{y}}_0$ base on $p_\theta(\mathbf{y}_0|\mathbf{y}_t, \mathbf{x})$ from the trained denoising network. Then, the noise for the $t-1$ timestep is added on the predicted $\hat{\mathbf{y}}_0$ based on the predefined $\mathbf{Q}$ to get the $\mathbf{y}_{t-1}$. Unlike the standard inference, we employ the easy-first strategy with the iterative denoising process. Since during training, we utilize the diffusion strategy and inject the information of the current timestep to the normalization, for the easy-first inference, we need to transcribe the iteration index of easy-first decoding to the corresponding timestep. For the iteration number $K$, if $K$ is larger than $T$, we reduce the iteration to the maximum diffusion steps $T$; and if $K$ is smaller than $T$, we map the current $k$ to timestep $t$,

\begin{equation}
t = \lfloor \frac{(K - k) * T}{K}\rfloor
\end{equation}

The noise for the timestep is not determined by $\mathbf{Q}$, but by the easy-first strategy, i.e., mask the characters with low recognition confidence as described in Section~\ref{Decoding}.

\section{Experiments}
\subsection{Datasets}
\subsubsection{English Datasets}
To evaluate the effectiveness and efficiency of our method, we conduct extensive experiments on almost all open-source datasets. We utilize the popular synthetic training dataset MJSynth~\cite{jaderberg2016reading} (MJ) and SynthText~\cite{gupta2016synthetic} (ST) as our synthetic training data. Since the importance of real datasets has been proved, we validate the performance of our models on real datasets collected by \cite{bautista2022scene}. Besides, we also train our models on the training set of the most recent benchmark dataset, Union14M-L~\cite{jiang2023revisiting}.

For English text recognition, We first evaluate our methods on the challenging dataset Union14M-Benchmark~\cite{jiang2023revisiting}. Other representative challenging datasets are also involved in the evaluation. Three large datasets: ArT~\cite{sun2019icdar} contains 35.1k curved and rotated hard-case images; COCO-Text~\cite{veit2016coco} comprises 9.8k occluded and distorted samples; Uber-Text~\cite{zhang2017uber} consists of 80.6k samples featuring vertical and rotated text. The occluded datasets~\cite{wang2021two} consists of images sheltered in weak (WOST) or heavy (HOST) degree. And the dataset WordArt~\cite{xie2022toward} contains images with texts in artistic style.

Besides, we provide the six widely used public benchmarks for reference but the models' performance on this benchmark is nearly saturated with some mislabeled images in it~\cite{jiang2023revisiting}. IIIT5K-Words~\cite{mishra2012scene} (IIIT5K) contains 5000 images collected from the website. There are 3000 images for testing, most of which are horizontal with high quality. Street View Text~\cite{wang2011end} (SVT) consists of 647 cropped word images from 249 street view images, which targets regular text recognition. ICDAR2013~\cite{karatzas2013icdar} (IC13) consists of 1095 regular-text images for testing. Two different versions are used for evaluation: 1015 and 857 images, which discard images that contain non-alphanumeric characters and less than three characters, respectively. ICDAR2015~\cite{karatzas2015icdar} (IC15) is a challenging dataset for recognition due to the degraded images collected without careful focusing, which contains 2077 cropped images. Some works used 1811 images for evaluation, discarding some distorted images.
SVT-Perspective~\cite{phan2013recognizing} (SVTP) comprises 645 images cropped from SVT, which is usually used for evaluating the performance of recognizing perspective text. CUTE80~\cite{risnumawan2014robust} (CUTE) consists of 288 curved images without lexicon, which are used for irregular text recognition.

\subsubsection{Chinese Datasets} \label{chinese_dataset}
To validate our models' performance on Chinese recognition, we train and evaluate our models on the Chinese benchmark dataset BCTR~\cite{chen2021benchmarking}, which consists of four subsets: scene, web, document, and handwriting, and each subset contains datasets for training, validation, and test. The scene dataset includes several public datasets, such as RCTW~\cite{shi2017icdar2017}, ReCTS~\cite{zhang2019icdar}, LSVT~\cite{sun2019icdar}, ArT~\cite{chng2019icdar2019} and CTW~\cite{yuan2019large}, with 112,471 samples for training, 14,059 samples for validation, and 14,059 samples for testing. The web dataset is collected from the MTWI~\cite{he2018icpr2018} dataset and contains 112,471, 14,059, and 14,059 samples for training, validation, and testing, respectively. The document subset is a synthetic dataset generated using Text Render in document style with 400,000 training samples, 50,000 validation samples, and 50,000 test samples. The handwriting subset is collected from a handwriting dataset SCUT-HCCDoc~\cite{zhang2020scut}, which consists of 74,603, 18,651, and 23,389 training, validation, and testing samples, respectively.

\subsection{Implementation Details}
\subsubsection{Model Settings}
For English recognition on most datasets, there are $37$ symbols covered for recognition, including numbers, case-insensitive characters, and $\langle {\rm EOS}\rangle$ token. For English recognition on Union14M, we employ the same charset with \cite{jiang2023revisiting} for fair comparison. We utilize the same charset with \cite{chen2021benchmarking} for Chinese recognition. We set the maximum length $L$ to $25$ for English and $40$ for Chinese recognition, respectively. The input image size is $32\times128$, and the patch size is $4\times8$. The details of the encoder and decoder transformer units settings are illustrated in Table~\ref{tab:transformer_set}, where $n$ is the number of Transformer units, $h$ is the number of attention heads, and $d_{model}$ is the dimension of the hidden layer in FFN. For English recognition, we utilize the DeiT-small configuration following \cite{bautista2022scene}, while for Chinese recognition with all three configurations. The number of easy-first iterations of the parallel decoder is set to $5$.

\begin{table}[!t]
    \caption{Configuration for small, base and large models. $d_{model}$ indicates the dimensions of the feature maps. $d_{MLP}$ represents the dimension of the intermediate features in the MLP layer. $h$ refers to the number of attention heads. $n$ represents the number of layers used in encoder and decoder.}
    \label{tab:transformer_set}
    \centering
    \begin{tabular}{ccccccc}
        \toprule
        \multirow{2}{*}{Models} & \multirow{2}{*}{$d_{model}$} & \multirow{2}{*}{$d_{MLP}$} & \multicolumn{2}{c}{Encoder} & \multicolumn{2}{c}{Decoder}  \\
        \cmidrule{4-5}
        \cmidrule{6-7}
         & & & $h$ & $n$ & $h$ & $n$ \\
        \midrule
        Small & 384 & 1536 & 6 & 12 & 12 & 1 \\
        Base & 768 & 3072 & 12 & 12 & 24 & 1  \\
        Large & 1024 & 4096 & 16 & 24 & 32 & 1\\
      \bottomrule
    \end{tabular}
\end{table}

\subsubsection{Model Training}
For English recognition, we train models on the synthetic (S), real (R), or Union14M-L datasets. For Chinese recognition, we combine the four subsets for model pre-training and then finetune the model on the specific subset. The input images are resized to $32\times128$ directly. We utilize Adam~\cite{kingma2015adam} as our optimizer with the 1cycle~\cite{smith2019super} learning rate scheduler. After $85\%$ of the total iterations, we replace it with Stochastic Weight Averaging (SWA)~\cite{izmailov2018avergaingwl}. A weight decay of $1e-4$ and a warmup ratio of $0.05$ are employed. Learning rates vary per model. 

We set the training epochs as 500 for Chinese recognition pre-training and 100 for specific subset finetuning. We train IPAD 50, 150, and 50 epochs for English recognition on synthetic, real, and Union14M-L datasets, respectively. The batch size is $384$, $192$, and $96$ for small, base, and large models, respectively. Data augmentations like Gaussian blur, Poisson noise, and rotation are randomly performed the same with PARSeq~\cite{bautista2022scene}.

\subsubsection{Model Evaluation}
We evaluate the English recognition for 36-charset word accuracy, including case-insensitive alphabets and digits. The 7248-sample average accuracy of six usual benchmarks is calculated with IC13-857 and IC15-1811, and the 7672-sample average accuracy is calculated with IC13-1015 and IC15-2077. For Chinese recognition evaluation, we post-process the predictions of the models following \cite{chen2021benchmarking}: (i) convert the full-width characters to half-width characters; (ii) convert all traditional Chinese characters to simplified characters; (iii) convert all English characters to lowercase; (iv) remove all spaces. 

\subsection{Comparisons with State-of-the-Art Methods}
\subsubsection{English Recognition}

\begin{table*}[t]
    \caption{Recognition results on Union14M-Benchmark. The average is calculated by directly averaging the seven results. MAERec-S$^*$ represents the model without pre-train. {PIMNet is reproduced with the ViT small encoder.}}
    \label{union}
    \centering
    \scalebox{0.8}{
    \begin{tabular}{lccccccccc}
    \toprule
    \multirow{2}{*}{Method} & \multirow{2}{*}{Curve} & Multi- & \multirow{2}{*}{Artistic} & \multirow{2}{*}{Contextless} & \multirow{2}{*}{Salient} & Multi- & \multirow{2}{*}{General} & \multirow{2}{*}{Avg} & Incomplete \\
    & & Oriented & & & & Words & & & ($\downarrow$)\\
    \midrule
    CRNN~\cite{shi2016end} & 19.4 & 4.5 & 34.2 & 44.0 & 16.7 & 35.7 & 60.4 & 30.7 & \textbf{0.9}\\
    SAR~\cite{li2019show} & 68.9 & 56.9 & 60.6 & 73.3 & 60.1 & \underline{74.6} & 76.0 & 67.2 & 2.1\\
    SATRN~\cite{wang2020decoupled} & 74.8 & 64.7 & 67.1 & 76.1 & 72.2 & 74.1 & 75.8 & 72.1 & \textbf{0.9}\\
    SRN~\cite{yu2020towards} & 49.7 & 20.0 & 50.7 & 61.0 & 43.9 & 51.5 & 62.7 & 48.5 & 2.2\\
    ABINet~\cite{fang2021read} & 75.0 & 61.5 & 65.3 & 71.1 & 72.9 & 59.1 & 79.4 & 69.2 & 2.6\\
    VisionLAN~\cite{wang2021two} & 70.7 & 57.2 & 56.7 & 63.8 & 67.6 & 47.3 & 74.2 & 62.5 & \underline{1.3}\\
    SVTR~\cite{du2022svtr} & 72.4 & 68.2 & 54.1 & 68.0 & 71.4 & 67.7 & 77.0 & 68.4 & 2.0\\
    MATRN~\cite{na2022multi} & \underline{80.5} & 64.7 & \underline{71.1} & 74.8 & \underline{79.4} & 67.6 & 77.9 & 74.6 & 1.7\\
    MAERec-S$^*$~\cite{jiang2023revisiting} & 75.4 & 66.5 & 66.0 & \underline{76.1} & 72.6 & \textbf{77.0} & \underline{80.8} & 73.5 & 3.5\\
    \midrule
    PIMNet~\cite{qiao2021pimnet} & 80.3 & \underline{79.8} & 68.4 & 75.9 & 77.8 & 68.3 & \underline{80.8} & \underline{75.9} & \underline{1.3} \\
    IPAD (Ours) & \textbf{85.2} & \textbf{83.3} & \textbf{72.1} & \textbf{78.4} & \textbf{81.4} & 73.7 & \textbf{82.2} & \textbf{79.5} & \underline{1.3}\\
    \bottomrule
    \end{tabular}
    }
\end{table*}

\begin{table}[!t]
    \centering
    \caption{Text recognition results on three large datasets. S and R represent the synthetic and real training dataset. $\sharp$ means that the model is reproduced by \cite{bautista2022scene}. {The subscripts $_A$ and $_N$ represent autoregressive and non-autoregressive version. PIMNet is reproduced with the ViT small encoder.}}
    \label{sota_largedatasets}
    \setlength{\tabcolsep}{1.5mm}{
    \begin{tabular}{lccccc}
    \toprule
     \multirow{2}{*}{Method} & Train & \multirow{2}{*}{ArT} & \multirow{2}{*}{COCO} & \multirow{2}{*}{Uber} & \multirow{2}{*}{Avg} \\
      & Data & & & & \\
    \midrule
     TRBA$^\sharp$~\cite{baek2019wrong} & S & 68.2 & 61.4 & 38.0 & 48.3\\
     PARSeq$_{A}$~\cite{bautista2022scene} & S & \textbf{70.7} & \underline{64.0} & 42.0 & \underline{51.8}\\
    \midrule
     ABINet$^\sharp$~\cite{fang2021read} & S & 65.4 & 57.1 & 34.9 & 45.2\\
    \midrule
    ViTSTR-S$^\sharp$~\cite{atienza2021vision} & S & 66.1 & 56.4 & 37.6 & 47.0\\
     PARSeq$_{N}$~\cite{bautista2022scene} & S & 69.1 & 60.2 & 39.9 & 49.7\\
     PIMNet~\cite{qiao2021pimnet} & S & 70.3 & 63.8 & \underline{42.3} & \underline{51.8}\\
     IPAD (Ours) & S & \underline{70.5} & \textbf{64.2} & \textbf{42.6} & \textbf{52.1}\\
    \midrule
    \midrule
     TRBA$^\sharp$~\cite{baek2019wrong} & R & 82.5 & 77.5 & 81.2 & 81.3\\
     PARSeq$_{A}$~\cite{bautista2022scene} & R & \textbf{84.5} & \textbf{79.8} & \textbf{84.5} & \textbf{84.1}\\
    \midrule
     ABINet$^\sharp$~\cite{fang2021read} & R & 81.2 & 76.4 & 71.5 & 74.6\\
    \midrule
     ViTSTR-S$^\sharp$~\cite{atienza2021vision} & R & 81.1 & 74.1 & 78.2 & 78.7 \\
     PARSeq$_{N}$~\cite{bautista2022scene} & R & 83.0 & 77.0 & 82.4 & 82.1\\
     PIMNet~\cite{qiao2021pimnet} & R & 83.7 & 78.6 & 83.2 & 83.0\\
     IPAD (Ours) & R & \underline{83.8} & \underline{78.8} & \underline{83.6} & \underline{83.3}\\
    \bottomrule
    \end{tabular}}
\end{table}

\begin{table}[!t]
    \centering
    \caption{Recognition results on occluded and artist images. {PIMNet is reproduced with the ViT small encoder.} All models are trained on synthetic datasets. }
    \setlength{\tabcolsep}{1.5mm}{
    \begin{tabular}{lccc}
        \toprule
        Model & HOST & WOST & WordArt \\
        \midrule
        VisionLAN~\cite{wang2021two} & 50.3 & 70.3 & - \\
        \midrule
        CornerTransformer~\cite{xie2022toward} & - & - & 70.8 \\
        \midrule
        PIMNet~\cite{qiao2021pimnet} & \underline{70.8} & \underline{81.5} & \underline{71.2}\\
        IPAD (Ours) & \textbf{71.7} & \textbf{82.2} & \textbf{71.4}\\
        \bottomrule
    \end{tabular}}
    \label{tab:ost_art}
\end{table}

\begin{table*}[t]
    \caption{Recognition results on six benchmark datasets. S$^1$ LM pretrained on WikiText-103~\cite{merity2017pointer}. MGP-STR$^\ddagger$ represents the small model for fair comparison. $\sharp$ means the model is reproduced by \cite{bautista2022scene}. {The subscripts $_A$ and $_N$ represent autoregressive and Non-autoregressive version. PIMNet is reproduced with the ViT small encoder.} The inference time is averaged on 7,672 images.}
    \centering
    \label{sota_comparison}
    \scalebox{0.7}{
    \begin{tabular}{clccccccccccccc}
        \toprule 
        & \multirow{2}{*}{Method} & Train & IIIT5K & SVT & \multicolumn{2}{c}{IC13} & \multicolumn{2}{c}{IC15} & SVTP & CUTE & \multicolumn{2}{c}{Average} & Params. & Time \\
         & & Data & 3000 & 647 & 857 & 1015 & 1811 & 2077 & 645 & 288 & 7248 & 7672 & (M)
         & (ms/img.)  \\
        \midrule
        \multirow{7}*{\rotatebox{90}{AR}} & ASTER~\cite{shi2018aster} & S & 93.4 & 89.5 & - & 91.8 & 76.1 & - & 78.5 & 79.5 & - & - & -
         & -\\
        & TRBA~\cite{baek2019wrong} & S & 87.9 & 87.5 & 93.6 & 92.3 & 77.6 & 71.8 & 79.2 & 74.0 & 84.6 & 82.8 & 49.6 
        & 16.7\\
        & SEED~\cite{qiao2020seed} & S & 93.8 & 89.6 & - & 92.8 & 80.0 & - & 81.2 & 83.6 & - & - & - 
        & -\\
        & RobustScanner~\cite{yue2020robustscanner} & S & 95.3 & 88.1 & - & 94.8 & - & 77.1 & 79.5 & 90.3 & - & 88.2 & - 
        & -\\
        & PTIE~\cite{tan2022pure} & S & 96.3 & \textbf{94.9} & - & \textbf{97.2} & \textbf{87.8} & \textbf{84.3} & \textbf{90.1} & 91.7 & - & \textbf{92.4} & 45.9 & 52.0\\
        & LevOCR~\cite{da2022levenshtein} & S$^1$ & 96.6 & 92.9 & 96.9 & - & 86.4 & - & 88.1 & 91.7 & 92.8 & - & 92.6
        & 60.5\\
        & PARSeq$_A$~\cite{bautista2022scene} & S & \textbf{97.0} & 93.6 & \textbf{97.0} & \underline{96.2} & 86.5 & 82.9 & 88.9 & \underline{92.2} & \underline{93.2} & \underline{91.9} & 23.8 
        & 17.5\\
        \midrule
        \multirow{8}*{\rotatebox{90}{NAR}}
        & ViTSTR-S~\cite{atienza2021vision} & S & 86.6 & 87.3 & 92.1 & 91.2 & 77.9 & 71.7 & 81.4 & 77.9 & 84.3 & 82.5 & 85.5 
        & 7.92\\ 
        & SRN~\cite{yu2020towards} & S & 94.8 & 91.5 & 95.5 & - & 82.7 & - & 85.1 & 87.8 & 90.4 & - & 54.7 & 12.1\\
        & VisionLAN~\cite{wang2021two} & S & 95.8 & 91.7 & 95.7 & - & 83.7 & - & 86.0 & 88.5 & 91.2 & - & 32.8 & 14.8\\
        & ABINet~\cite{fang2021read} & S$^1$ & 96.2 & 93.5 & \textbf{97.4} & 95.7 & 86.0 & 85.1 & 89.3 & 89.2 & 92.6 & - & 36.7 & 23.4\\
        & PARSeq$_N$~\cite{bautista2022scene} & S & 95.7 & 92.6 & 96.3 & 95.5 & 85.1 & 81.4 & 87.9 & 91.4 & 92.0 & 90.7 & 23.8 & 11.7\\
        & MGP-STR$^\ddagger$~\cite{wang2022multi} & S & 95.3 & 93.5 & 96.4 & - & 86.1 & - & 87.3 & 87.9 & 92.0 & - & 52.6 & 9.37\\
        & PIMNet~\cite{qiao2021pimnet} & S & 96.6 & 93.2 & 96.6 & 95.7 & 86.0 & 82.3 & 87.8 & 91.0 & 92.6 & 91.4 & 24.2 & 14.6\\
        & IPAD (Ours) & S & \underline{96.8} & \underline{94.3} & \underline{97.0} & 95.6 & \underline{86.7} & \underline{83.0} & \underline{89.3} & \textbf{93.4} & \textbf{93.3} & \underline{91.9} & 24.7 & 15.0\\
        \midrule
        \midrule
        \multirow{2}*{\rotatebox{90}{AR}} & TRBA$^\sharp$~\cite{baek2019wrong} & R & 98.6 & 97.0 & 97.6 & 97.6 & 89.8 & 88.7 & 93.7 & 97.7 & 95.7 & 95.2 & 49.8 & 21.5\\
        & PARSeq$_{A}$~\cite{bautista2022scene} & R & \textbf{99.1} & \textbf{97.9} & \underline{98.3} & \textbf{98.4} & \underline{90.7} & \underline{89.6} & \textbf{95.7} & \textbf{98.3} & \textbf{96.4} & \textbf{96.0} & 23.8 & 17.5\\
        \midrule
        \multirow{4}*{\rotatebox{90}{NAR}} & ViTSTR-S$^\sharp$~\cite{atienza2021vision} & R & 98.1 & 95.8 & 97.6 & 97.7 & 88.4 & 87.1 & 91.4 & 96.1 & 94.7 & 94.3 & 21.4 & 6.94\\
        & ABINet$^\sharp$~\cite{fang2021read} & R & 98.6 & 97.8 & 98.0 & 97.8 & 90.2 & 88.5 & 93.9 & 97.7 & 95.9 & 95.2 & 36.9 & 22.8 \\ 
        & PARSeq$_{N}$~\cite{bautista2022scene} & R & 98.3 & 97.5 & 98.0 & \underline{98.1} & 89.6 & 88.4 & 94.6 & 97.7 & 95.7 & 95.2 & 23.8 
        & 11.7\\
        & PIMNet~\cite{qiao2021pimnet} & R & 98.9 & \underline{97.8} & \textbf{98.4} & \underline{98.1} & 89.6 & 88.4 & 94.3 & \underline{97.9} & 95.9 & 95.4 & 24.2 & 14.6\\
        & IPAD (Ours) & R & \underline{99.0} & 97.7 & 98.1 & \underline{98.1} &  \textbf{90.8} & \textbf{89.8} & \underline{95.5} & \underline{97.9} & \textbf{96.4} & \textbf{96.0} & 24.7
        & 15.0\\
        \bottomrule
    \end{tabular}}
\end{table*}

Given recent studies highlighting performance saturation on six usual benchmarks, we first conduct experiments on the Union14M-Benchmark to demonstrate the performance of our model. As shown in Table~\ref{union}, trained with the Union14M-L, our IPAD improves the accuracy by an average of 4.9\%. Notably, on the curved, multi-oriented, and salient datasets, the IPAD surpasses the previous best model by 4.7\%, 15.1\%, and 2.0\%, respectively, which demonstrates the assistance of abundant contextual information. However, leveraging contextual information during recognition also has drawbacks, such as the unsatisfactory performance on the incomplete dataset. As shown in Table~\ref{union}, the best model on the incomplete dataset is CRNN~\cite{shi2016end} and SATRN~\cite{wang2020decoupled}, which do not leverage linguistic knowledge. 
This occurs because models that leverage contextual information are trained on large volumes of complete English words. Incomplete words fall outside their learned linguistic patterns, causing these context-dependent models to fill in missing characters in an attempt to align the results with their internal linguistic knowledge. For example, when faced with an incomplete word “convenienc”, the model is likely to add an “e” to form “convenience”, as this fits the contextual information they get during training.

Trained with synthetic or real datasets, we evaluate our model on large and intricate datasets. These include the three extensive datasets presented in Table~\ref{sota_largedatasets} and the occluded and artistic datasets featured in Table~\ref{tab:ost_art}. As evidenced in Table~\ref{sota_largedatasets}, when trained on synthetic datasets, our IPAD exhibits superior performance to the {autoregressive (AR) version} PARSeq$_A$. While models trained on real datasets may not surpass the AR model in terms of performance, they excel in reduced inference time. Compared with other {non-autoregressive (NAR)} models, our models manifest enhanced performance, with IPAD showing improvements of $1.2\%$. The results in Table~\ref{tab:ost_art} suggest that our models excel in recognizing text on occluded images due to their advanced context learning capabilities. Additionally, they surpass the CornerTransformer~\cite{xie2022toward} in artistic text recognition.

Following previous works, We also compare our model IPAD with other published state-of-the-art methods on the six widely used benchmarks. As delineated in Table~\ref{sota_comparison}, it is evident that AR methods typically outperform their NAR counterparts. Even though an external language module can function as a recognition checker and refine the preliminary recognition results, the performance of LM methods consistently lags behind that of the AR models. 
Furthermore, by leveraging the diffusion model, IPAD elevates performance across nearly all datasets, marking a notable impact on irregular ones. This paper tries to balance the accuracy and the efficiency. 
Our IPAD demonstrates accuracy comparable to most autoregressive methods while ensuring accelerated inference speed. Compared to the AR-based method PARSeq 1 refinement~\cite{bautista2022scene}, our methodology produces similar accuracy on average but needs 2.5ms less time. 
Aligning with PARSeq~\cite{bautista2022scene}, we also train our models utilizing real data. As delineated in Table~\ref{sota_comparison}, trained on real datasets, our IPAD can also align in accuracy with the refined autoregressive model, PARSeq$_A$.

\subsubsection{Chinese Recognition}
\begin{table*}[!t]
\caption{Chinese text recognition results on the BCTR datasets. The `Combined' dataset contains about 1.1M data.}
\label{tab:chinese_sota}
\centering
\scalebox{0.8}{
\begin{tabular}{lcccccccc}
\midrule
Method & (Pre)Training Data & Finetuning Data & Scene & Web & Doc. & Handw. & Avg & Params. (M)\\
\midrule
CRNN~\cite{shi2016end} & Specific & - & 54.9 & 56.2 & 97.4 & 48.0 & 68.0 & 12\\
ASTER~\cite{shi2018aster} & Specific & - & 59.4 & 57.8 & 97.6 & 45.9 & 69.8 & 27 \\
MORAN~\cite{luo2019moran} & Specific & - & 54.7 & 49.6 & 91.7 & 30.2 & 62.7 & 29\\
SAR~\cite{li2019show} & Specific & - & 53.8 & 50.5 & 96.2 & 31.0 & 64.0 & 28\\
SEED~\cite{qiao2020seed} & Specific & - & 45.4 & 31.4 & 96.1 & 21.1 & 57.1 & 36\\
MASTER~\cite{lu2021master} & Specific & - & 62.1 & 53.4 & 82.7 & 18.5 & 61.4 & 63\\
ABINet~\cite{fang2021read} & Specific & - & 60.9 & 51.1 & 91.7 & 13.8 & 62.9 & 53\\
TransOCR~\cite{chen2021scene} & Specific & - & 67.8 & 62.7 & 97.9 & \textbf{51.7} & 74.8 & 84\\
PIMNet$_{\text{Small}}$~\cite{qiao2021pimnet} & Specific & - & 78.8 & 67.8 & 98.4 & 37.6 & 77.9 & 30\\
IPAD$_{\text{Small}}$(Ours) & Specific & - & \textbf{79.9} & \textbf{69.8} & \textbf{98.8} & 45.4 & \textbf{79.9} & 31\\
\midrule
MaskOCR$_{\text{ViT-S}}$~\cite{lyu2022maskocr} & \multirow{3}*{\thead{100M Unlabled Real\\+ 100M Synthetic}}& \multirow{3}{*}{Specific} & 71.4 & 72.5 & 98.8 & 55.6 & 78.1 & 36 \\
MaskOCR$_{\text{ViT-B}}$~\cite{lyu2022maskocr} & & & 73.9 & 74.8 & 99.3 & 63.7 & 80.8 & 100 \\
MaskOCR$_{\text{ViT-L}}$~\cite{lyu2022maskocr} & & & 76.2 & 76.8 & \textbf{99.4} & \textbf{67.9} & 82.6 & 318 \\
\midrule
PIMNet$_{\text{Small}}$~\cite{qiao2021pimnet} & \multirow{4}{*}{Combined} & \multirow{4}{*}{Specific} & 81.0 & 80.0 & 98.9 & 59.0 & 83.4 & 30\\
IPAD$_{\text{Small}}$(Ours) & & & 82.0 & 80.9 & 99.1 & 61.7 & 84.4 & 31\\
IPAD$_{\text{Base}}$(Ours) & & & 83.5 & 81.9 & 99.2 & 63.0 & 85.4 & 110 \\
IPAD$_{\text{Large}}$(Ours) & & & \textbf{84.2} & \textbf{82.6} & 99.2 & 64.2 & \textbf{85.9} & 342\\
\midrule
\end{tabular}}
\end{table*}

To elucidate the capabilities of our model in more complex, non-Latin character recognition scenarios, we compare IPAD against the state-of-the-art Chinese recognition models on the BCTR dataset. 
As shown in Table~\ref{tab:chinese_sota}, our IPAD without pre-training has surpassed the preceding methods on three subsets. Specifically, it has introduced improvements of $12.1\%$ on the scene subset, $7.1\%$ on the web subset, $0.9\%$ on the document subset, and an average enhancement of $4.8\%$. 

Moreover, we initially pre-train the IPAD on the combined BCTR training dataset before finetuning it on specific subsets since the dataset for Chinese recognition is relatively small, and the STR models still demand large datasets. This approach produces notable improvements, and our pretrained models demonstrate their superior performance. Notably, IPAD$_{\text{Small}}$ manifests substantial advancements in web and handwriting text recognition with improvements of $11.1\%$ and $18.0\%$, respectively. 
We attribute part of MaskOCR~\cite{lyu2022maskocr}'s superior performance in handwriting recognition to its reliance on extensive pretraining datasets, potentially including a wealth of handwriting samples and contextually analogous texts. However, the marked proficiency of our model in scene and web subsets underscores its ability to perceive linguistic information. Besides, the pronounced efficacy of our model in a smaller configuration is particularly notable ($10.6\%$ and $8.4\%$ improvements compared with MaskOCR$_{\text{ViT-S}}$ on scene and web subsets, respectively), demonstrating its practical utility and efficiency.

\subsection{Ablation Studies}
In this subsection, we conduct ablation studies to compare IPAD with our conference version PIMNet~\cite{qiao2021pimnet} and show the effectiveness of each proposed module and strategy within both English and Chinese recognition contexts. All experiments within this section utilize small models to maintain consistency and focus on component-specific impacts. In PIMNet, we utilize the mimicking learning strategy to enhance the training of the parallel decoder, where the FFN outputs of the parallel decoder are aligned to the FFN feature of an autoregressive decoder with the same encoder. Additionally, to maintain consistency in the encoder, we replace the CNN encoder from the conference version with the same small ViT encoder used in IPAD.

\subsubsection{Analysis of Different Components} 

\begin{table*}[!t]
\caption{Experimental results using different components. Recognition accuracy on English is the average accuracy over the 7248 samples. The Chinese recognition trains on the combined BCTR dataset without finetuning. $_\text{w/o}$ means without any mimicking or diffusion step.}
\centering
\scalebox{0.9}{
\begin{tabular}{lccccccc}
\toprule
\multirow{2}{*}{Model} & \multirow{2}{*}{Base} & \multirow{2}{*}{Easy-first} & \multirow{2}{*}{Mimicking} & \multirow{2}{*}{Diffusion}& \multicolumn{2}{c}{English} & \multirow{2}{*}{Chinese} \\
 & & & & & Synthetic & Real & \\
\midrule
Baseline & \ding{51} & \ding{55} & \ding{55} & \ding{55} & 92.00 &  95.35 & 80.56\\
IPAD$_\text{w/o}$ & \ding{51} & \ding{51} & \ding{55} & \ding{55} & 92.47 & 95.85 & 82.78\\
PIMNet & \ding{51} & \ding{51} & \ding{51} & \ding{55} & 92.63 & 95.94 & 82.86\\
IPAD & \ding{51} & \ding{51} & \ding{55} & \ding{51} & \textbf{93.25} & \textbf{96.39} & \textbf{84.08}\\
\bottomrule
\end{tabular}
}
\label{tab:components}
\end{table*}

We conduct ablation studies on different components to dissect their contribution towards performance improvement, as detailed in Table~\ref{tab:components}. The foundational baseline consists of a purely non-autoregressive model with our parallel decoder. Subsequently, for IPAD$_{\text{w/o}}$ (without any mimicking or diffusion step), we implement the easy-first decoding strategy, utilizing linguistic information throughout the iterative decoding process and formulating predictions for the masked characters based on their relationships with the unmasked ones. As shown in Table~\ref{tab:components}, the deployment of easy-first decoding exhibits discernible improvements across both English and Chinese datasets, highlighting a notable $2.22\%$ improvement on the Chinese datasets.  

The diffusion strategy empowers the IPAD with a profound implicit understanding of contextual relationships within words by incorporating a time-dependent noising and denoising generation process during the training phase. It could explore more possible combinations of context. Therefore, the IPAD performs better compared with IPAD$_{\text{w/o}}$, leading to advancements of $0.78\%$ and $0.54\%$ on English trained with synthetic and real datasets, and $1.30\%$ on Chinese, respectively. The diffusion strategy's effectiveness demonstrates its pivotal role in enabling the model to navigate and comprehend the intricate linguistic landscapes and relationships inherent in text recognition tasks. In contrast, incorporating the mimicking learning strategy used in our conference version only enables PIMNet marginally superior performance relative to IPAD$_{\text{w/o}}$. The increments are almost negligible, especially for models trained with real datasets for English and Chinese recognition. 

\subsubsection{Comparison between Mimicking Learning and Discrete Diffusion}
\begin{figure}[!t]
\centering
\subfloat[Normalized cosine similarities of FFN outputs for relatively simple English test images.]
{\includegraphics[width=1\columnwidth]{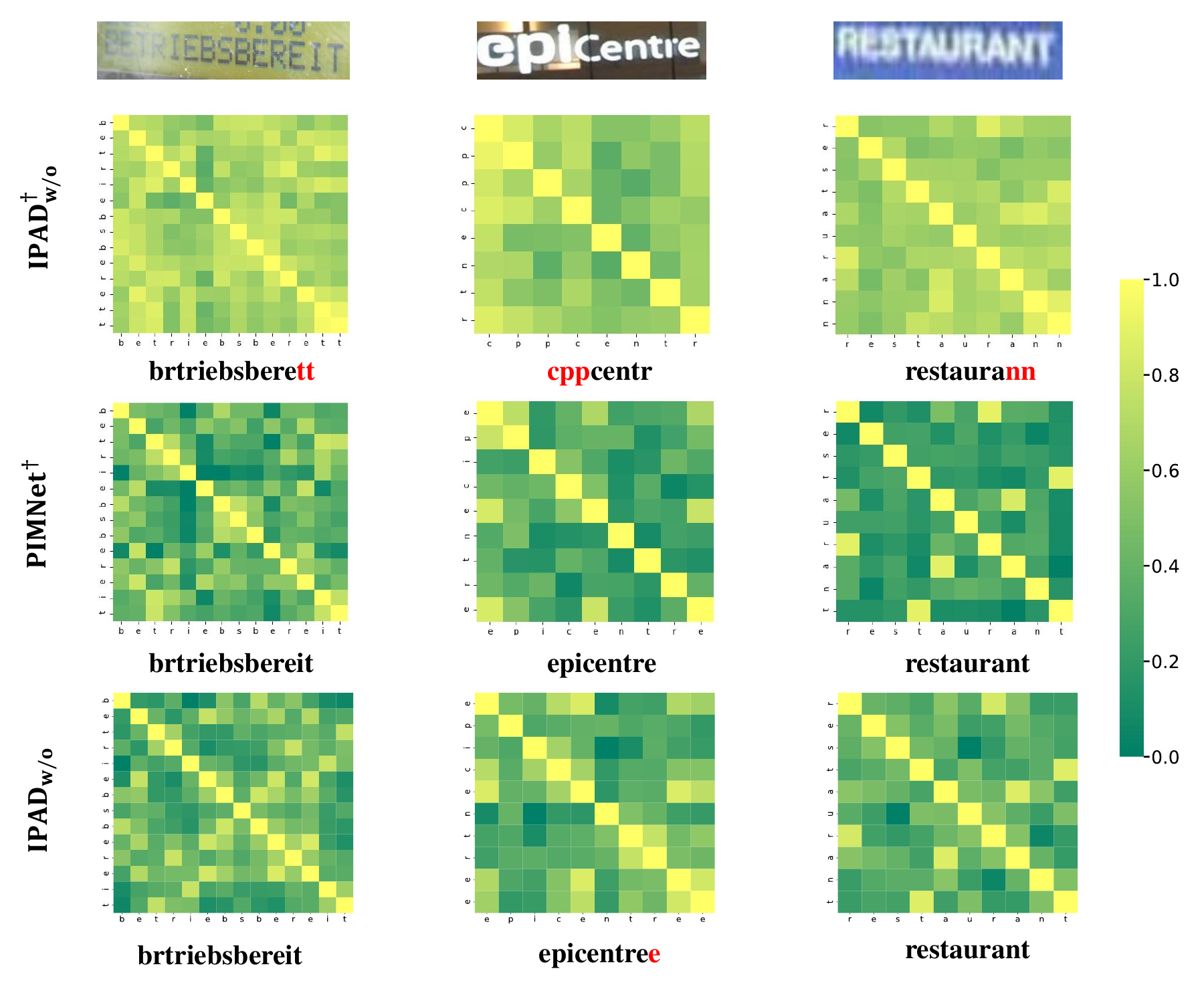}%
\label{ffnmap_origin}}
\hfil
\subfloat[Normalized cosine similarities of FFN outputs for more challenging English and Chinese test images.]
{\includegraphics[width=1\columnwidth]{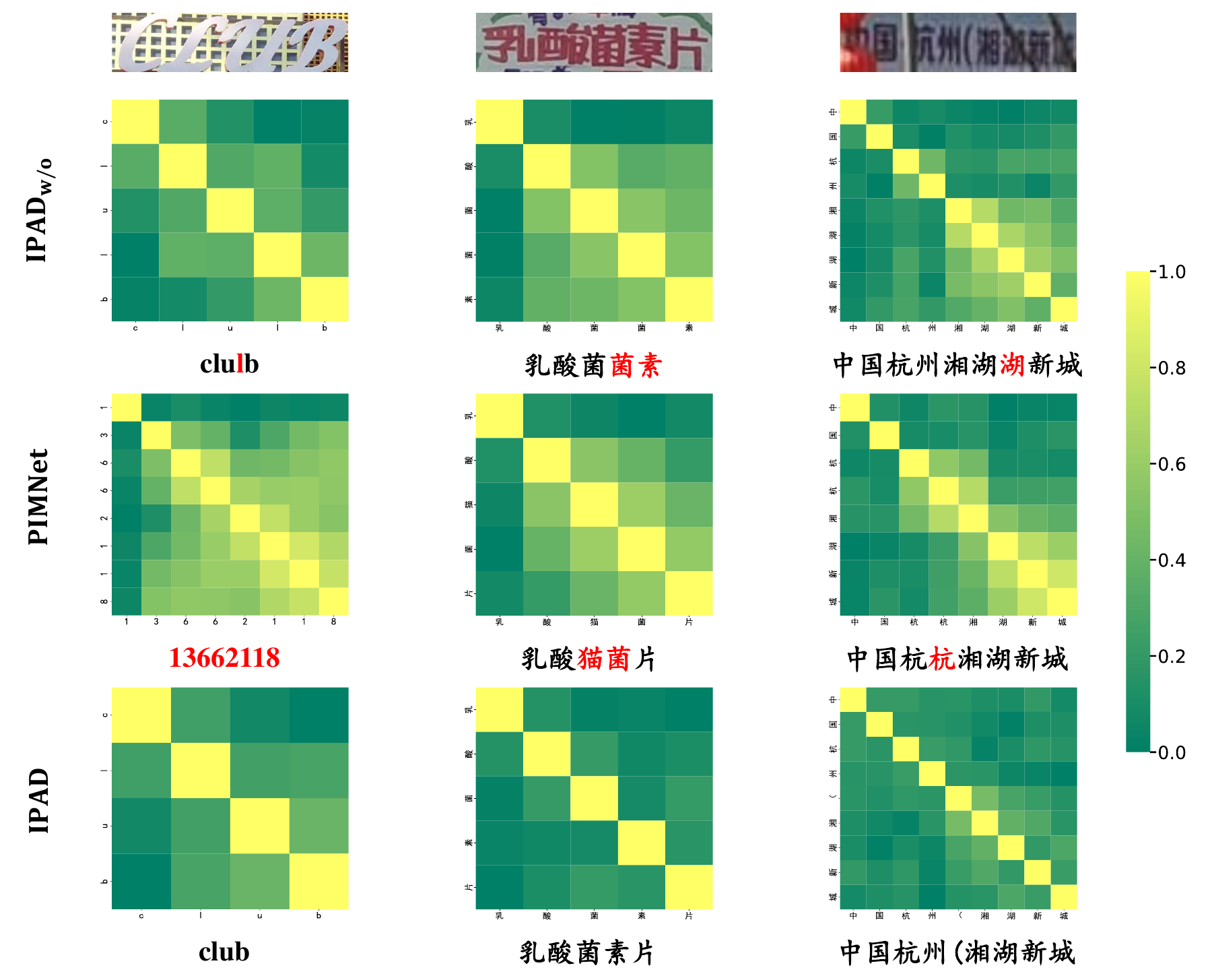}%
\label{ffnmap_new}}
\caption{Comparison of normalized cosine similarities of FFN outputs across different encoders. $^\dagger$ indicates models using a CNN encoder. $_\text{w/o}$ denotes models without mimicking or diffusion steps.}
\label{ffnmap_compare}
\end{figure}

To analyze and compare the efficacy of the mimicking learning and discrete diffusion strategies further, we provide visual representations of the cosine similarities of FFN outputs. As depicted in Fig.~\ref{ffnmap_new}, the parallel decoder in the early iterations tends to predict similar outputs due to the fully parallel decoding, thus the similarities among neighboring positions tend to be large. The similar outputs of FFN may mislead the final predictions, which tend to be duplicated and wrong. In PIMNet, erroneous predictions of preceding characters can influence the recognition of subsequent characters and introduce noise into the FFN outputs. This occurs as it mimics the FFN feature of the autoregressive decoder, which is inherently incapable of learning bidirectional context and is prone to significant error accumulation. Thus, the similarities across positions in the left part are relatively large, which may mislead the final predictions.

Conversely, the application of the diffusion strategy allows IPAD to yield more distinguishable FFN outputs in challenging cases, for instance, against complex backgrounds, and in cases of occluded and irregular text, as demonstrated in Fig.~\ref{ffnmap_new}. This substantiates the diffusion strategy's role in enhancing the model's adaptability in deciphering text under intricate conditions.

\subsubsection{Analysis of Encoder}
\begin{table*}[]
    \centering
    \caption{Ablation study on different encoder. IC13 here is IC13-857 and IC15 is IC1-1811. $^\dagger$ indicates the model uses a CNN encoder. $_\text{w/o}$ means without any mimicking or diffusion step.}
    \scalebox{0.9}{
    \begin{tabular}{lcccccccc}
    \toprule
    Model & encoder & IIIT5K & SVT & IC13 & IC15 & SVTP & CUTE & Avg.\\
    \midrule
    IPAD$^\dagger_{\text{w/o}}$ & CNN & 94.9 & 90.1 & 94.7 & 82.9 & 83.4 & 86.1 & 90.1\\
    PIMNet$^\dagger$ & CNN & 95.2 & 91.2 & 95.2 & 83.5 & 84.3 & 84.4 & 90.5 \\
    \midrule
    IPAD$_{\text{w/o}}$ & ViT & 96.7 & 94.0 & 96.4 & 85.0 & 87.6 & 91.0 & 92.5 \\
    PIMNet & ViT & 96.6 & 93.2 & 96.6 & 86.0 & 87.8 & 91.0 & 92.6\\
    \bottomrule
    \end{tabular}}
    \label{tab:encoder}
\end{table*}

We analyze the influence of different encoders to gain insights into their impacts on performance. In IPAD$^\dagger_{\text{w/o}}$ and PIMNet$^\dagger$, we employ a Feature Pyramid Network (FPN) coupled with a ResNet-50, supplemented by two additional transformer units as the encoder. In IPAD$_{\text{w/o}}$ and PIMNet, a Vision Transformer (ViT) with $4\times8$ image patches is deployed.

As revealed in Table~\ref{tab:encoder}, leveraging ViT improves the performance of PIMNet on the six standard English recognition benchmarks markedly, introducing improvements exceeding $2\%$. 
Furthermore, because of the refined quality of image features encoded by ViT, the model exhibits diminished reliance on the decoder. Consequently, the mimicking learning strategy utilized in our conference paper yields marginal gains of $0.1\%$ with the ViT encoder, contrasting the $0.4\%$ gains observed with the previous CNN encoder. This phenomenon indicates that the mimicking learning strategy in our conference paper is somewhat inadequate to enhance the parallel decoder’s contextual information learning capabilities in the context of a ViT encoder.

In Fig.~\ref{ffnmap_origin}, the CNN encoder shows limited capacity in capturing relationships among detailed image features, which is why mimicking learning helps PIMNet$^\dagger$ produce more distinguishable FFN outputs and improve prediction accuracy. However, when switching to the ViT encoder, which synthesizes higher-quality visual features and better captures visual relationships, the need for mimicking learning decreases, and its limitations become more apparent. {The ViT encoder enables IPAD$_{\text{w/o}}$ to generate clearer and more discriminative FFN features on the same test images as IPAD$^\dagger_{\text{w/o}}$. Notably, IPAD$_{\text{w/o}}$ is able to recognize these relatively simple test images in Fig.\ref{ffnmap_origin} almost correctly without any specific decoding design. To further illustrate the advantages of IPAD, Fig.\ref{ffnmap_new} presents more challenging English and Chinese test images.
As shown in Fig.~\ref{ffnmap_new}, the FFN pattern for IPAD$_{\text{w/o}}$ is also distinct enough.} The second row of Fig.~\ref{ffnmap_new} shows the FFN outputs of PIMNet with the ViT encoder. By comparing this with the first row, it becomes clear that early recognition errors (such as those seen with the ``club" figure in the mimicking learning PIMNet) negatively impact the discriminability of subsequent predictions. This occurs because mimicking learning attempts to replicate the autoregressive decoding process, which accumulates errors from earlier stages of decoding. 

{In summary, Table~\ref{tab:encoder} demonstrates the numerical improvements gained from the ViT encoder. Meanwhile, Fig.\ref{ffnmap_origin} compares the FFN patterns of both CNN-based and ViT-based models. The CNN-based IPAD$^\dagger_{\text{w/o}}$ exhibits less clarity, whereas the ViT-based IPAD$_{\text{w/o}}$ generates more distinct and discriminative FFN features, even without specific decoding strategies. Fig.\ref{ffnmap_origin} further highlights the performance on more challenging English and Chinese test images, showing that the ViT encoder effectively handles complex visual features and improves feature representation, reinforcing the advantages of our proposed method.}

\subsubsection{Effect of Timestep Setting in Discrete Diffusion}
The diffusion model's performance is intricately linked to how noise samples are generated, and this process is significantly influenced by the specific timestep schedule. How to set the timestep embedding plays a pivotal role in the model's denoising capabilities. In the context of the presented results in Table~\ref{tab:timestep}, it becomes evident that if the model ignores the timestep information during noising and denoising, its performance will suffer greatly. This decline in accuracy is more obvious for Chinese recognition since the linguistic context in Chinese is more complex. The phenomenon underscores the significance of timestep information in equipping the model to select the most pertinent contextual cues effectively. When Adaptive Layer Normalization is utilized with a learnable timestep embedding, the accuracy is also less than satisfactory compared with the sinusoidal positional embedding's results. The model demonstrates commendable results when provided with a little timestep than the length of the text. For example, although the prescribed text length for Chinese is $40$, setting the timestep to $T=25$ seems optimal for Chinese recognition. This preference can be explained by the fact that most Chinese text images contain fewer than 25 characters.

\begin{table}[]
    \centering
    \caption{The influences of different ways to set the timestep embedding. W/o Timestep means that we do not indicate the timestep during the denoising process.}
    \label{tab:timestep}
    \setlength{\tabcolsep}{1mm}{
    \begin{tabular}{lccc}
    \toprule
    \multirow{2}{*}{Method} & \multicolumn{2}{c}{English} & \multirow{2}{*}{Chinese}\\
    & Synth & Real & \\
    \midrule
    W/o Timestep & 92.89 & 96.03 & 81.97 \\
    Learnable Embedding & 92.95 & 96.21 & 82.65\\
    \midrule
    Sinusoidal Embedding (T=10) & 93.16 & 96.32 & 83.58\\
    Sinusoidal Embedding (T=25) & \textbf{93.25} & \textbf{96.39} & \textbf{84.08}\\
    Sinusoidal Embedding (T=40) & 93.02 & 96.14 & 83.99\\
    Sinusoidal Embedding (T=80) & 93.10 & 95.93 & 83.96\\
    \bottomrule
    \end{tabular}}
\end{table}

\subsubsection{Analysis of Iteration Number} 
The number of iterations is an important hyper-parameter in our method, so we conduct experiments to analyze the effect caused by the iterations number. As shown in Table~\ref{tab:eng_iter} and Table~\ref{tab:chinese_iteration}, the fully parallel IPAD model with only one iteration works poorly on both English and Chinese recognition. As we discussed previously, the fully parallel decoding lacks useful context information, which impacts the optimization procedure. When an additional iteration is adopted, the performance is improved significantly, especially on Scene (from $77.17\%$ to $80.53\%$), Web (from $77.61\%$ to $79.72\%$), and Handwriting (from $50.82\%$ to $57.52\%$) subsets of BCTR. With the increment of iterations, the accuracy and inference time increase accordingly. Five is selected as the number of iterations in the implementation to achieve a balance. Note that the timestep $T$ of diffusion in our implementation is $25$, so the model with $25$ iterations is the best. By utilizing different numbers of iterations, we can achieve various goals, i.e., achieve high performance or speed, which also shows the flexibility of our model.

\begin{table}[!t]
\caption{The comparison of accuracy and inference time with different iterations.}
\label{tab:eng_iter}
\centering
\setlength{\tabcolsep}{1.2mm}{
\begin{tabular}{ccccccccc}
\toprule
Iter & 1 & 2 & 3 & 5 & 10 & 15 & 25\\
\midrule
S & 92.15 & 92.98 & 92.96 & 93.25 & 93.41 & 93.46 & \textbf{93.60}\\
R & 94.90 & 95.94 & 96.27 & \textbf{96.39} & 96.37 & 96.36 & 96.27\\
\midrule
Time & 9.39 & 11.3 & 12.9 & 15.0 & 23.9 & 32.4 & 47.8\\
\bottomrule
\end{tabular}}
\end{table}

\begin{table}[!t]
\caption{The comparison of accuracy and inference time with different iterations for Chinese recognition.}
\label{tab:chinese_iteration}
\centering
\setlength{\tabcolsep}{1.6mm}{
\begin{tabular}{lcccccc}
\toprule
\multirow{2}*{Iter} & \multirow{2}*{Scene} & \multirow{2}*{Web} & \multirow{2}*{Doc} & \multirow{2}*{Hand} & \multirow{2}*{Avg.} & Time 
\\
& & & &&&(ms/img.)\\
\midrule
1 & 77.17 & 77.61 & 98.66 & 50.82 & 80.23 & 9.45\\
2 & 80.56 & 79.68 & 98.94 & 57.51 & 82.98 & 11.6\\
3 & 81.30 & 80.15 & 99.01 & 59.24 & 83.63 & 13.9 \\
5 & 81.85 & 80.38 & 99.00 & 60.30 & 84.04 & 16.8\\
10 & 82.18 & 80.49 & 99.04 & 60.84 & 84.28 & 26.0\\
15 & 82.25 & 80.56 & 99.06 & 60.99 & 84.35 & 34.7\\
25 & 82.23 & 80.50 & 99.06 & 61.07 & 84.35 & 51.6\\
\bottomrule
\end{tabular}}
\end{table}

\subsubsection{Training Procedure for Chinese Recognition}
\begin{table}[!t]
\caption{Text recognition results on the BCTR datasets. $_{\text{w/o}}$ means without any mimicking or diffusion step. `Spec.' means that we utilize the corresponding subset to train a specific model for the four datasets, while `Comb.' means that we combine the four subsets. FT indictaes whether the finetuning is employed. For finetuning, the model is finetuned on specific dataset.}
\label{tab:chinese_training}
\centering
\setlength{\tabcolsep}{1.6mm}{
\begin{tabular}{ccccccc}
\toprule
Model & Data & FT & Scene & Web & Doc. & Handw.\\
\midrule
\multirow{3}{*}{IPAD$_{\text{w/o}}$} & Spec. & \ding{55} & 78.62 & 66.44 & 98.07 & 33.90\\
 & Comb. & \ding{55} & 80.71 & 79.06 & 98.27 & 57.14\\
 & Comb. & \ding{51} & 81.10 & 79.81 & 98.63 & 58.00\\
\midrule
\multirow{3}{*}{IPAD} & Spec. & \ding{55} & 79.94 & 69.75 & 98.79 & 43.74\\
& Comb. & \ding{55} & 81.85 & 80.38 & 99.00 & 60.30\\
& Comb. & \ding{51} & 82.02 & 80.89 & 99.11 & 61.70\\
\bottomrule
\end{tabular}}
\end{table}

As described in section~\ref{chinese_dataset}, the BCTR dataset consists of four subsets. Previous methods usually train and test on each specific subset. And some methods utilize synthetic datasets to pretrain their models and then finetune them on different subsets. However, we notice that combining the four subsets for training and directly testing the model on the particular task can outperform training on the specific dataset, especially on the Web and Handwriting dataset, as shown in Table~\ref{tab:chinese_training}. And the validation set is also the combination of the four sub-datasets. We argue that the reason is the small amount of data because combining all four datasets can only result in 1,096,238 samples for training, whose amount is one fourteenth of the English synthetic dataset and about a third of the English real dataset. Besides, the Chinese charset is much larger than the English, which demands more samples to perform similarly. If we further finetune the model trained on the whole dataset on a particular sub-dataset, we could improve the model's performance further.


\subsubsection{Analysis of Inference Speed}
\begin{table}[!t]
\caption{The comparison of inference time among different decoding strategies. $_\text{w/o}$ means without any mimicking or diffusion step. We evaluated the model, trained on synthetic datasets, using the six English benchmarks to maintain consistency with previous studies.}
\label{tab:inference_time}
\centering
\begin{tabular}{lcc}
\toprule
Methods & Time(ms/image) & Acc.\\
\midrule
CTC based & 8.14 & 90.3\\
Attention based & 43.9 & 92.0\\
\midrule
IPAD$_{\text{w/o}}$ (1 iteration) & 9.39 & 90.8\\
IPAD$_{\text{w/o}}$ (5 iterations) & 14.6 & 92.5\\
\midrule
IPAD (1 iteration) & 9.39 & 92.2\\
IPAD (5 iteration) & 15.0 & 93.3\\
\bottomrule
\end{tabular}
\end{table}

\begin{figure*}[!ht]
\centering
\includegraphics[width=0.8\textwidth]{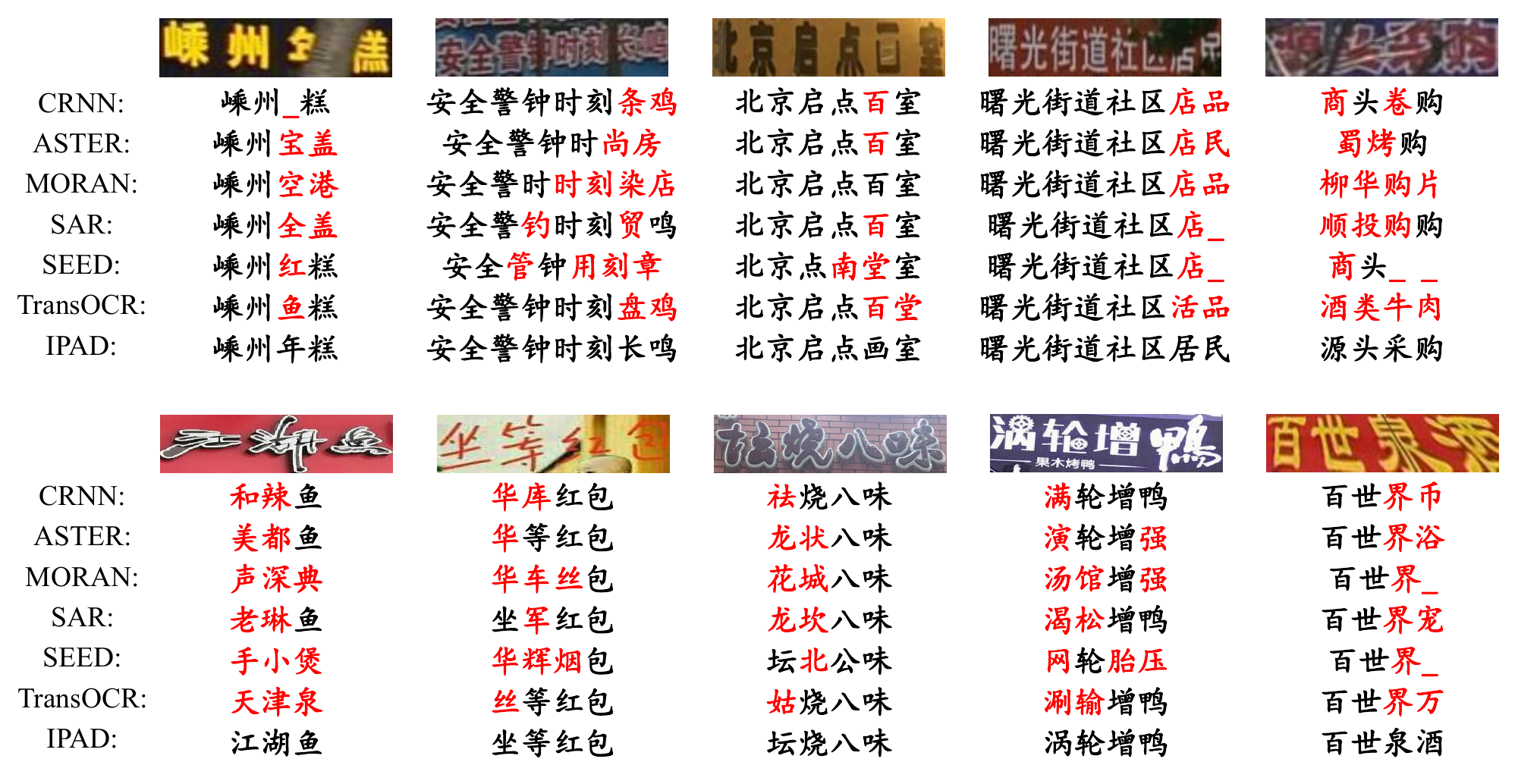}
\caption{Comparison of recognition results on the scene subset of the BCTR dataset. The characters in red are wrongly recognized. The red ``\_" means missing characters.}
\label{fig:chinese_samples}
\end{figure*}

\begin{figure}[!ht]
\centering
\includegraphics[width=1\columnwidth]{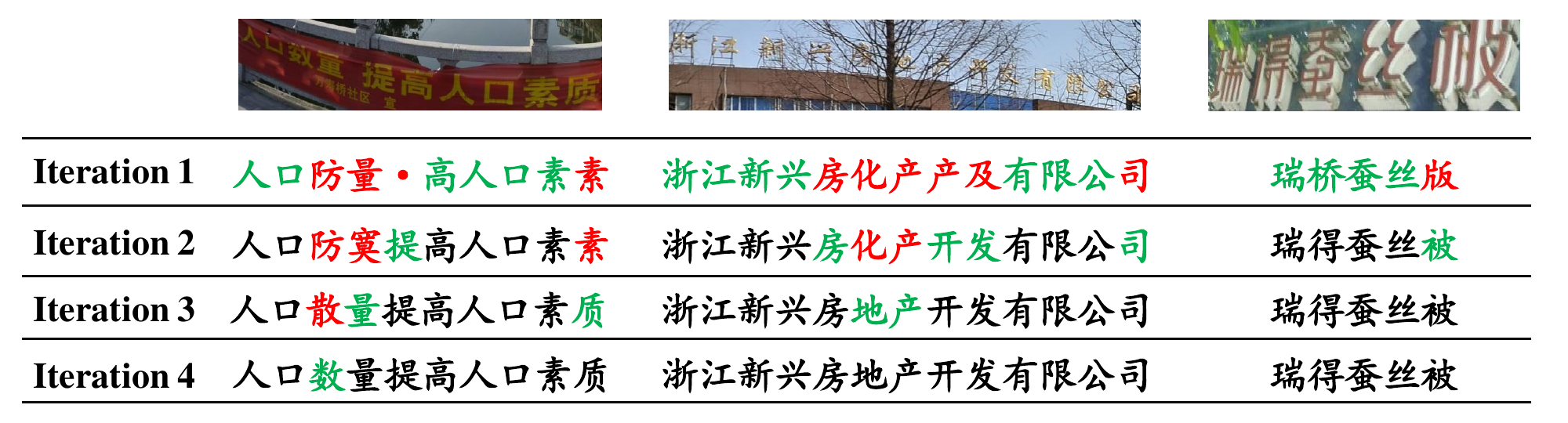}
\caption{Some examples to illustrate the iterative generation of IPAD with easy first.}
\label{fig:chinese_iter}
\end{figure}

To further verify our advantage in efficiency, we compare our method with CTC and AR attention-based decoders. To remove the influence of the encoder, we evaluate all methods with the same encoder as our IPAD. As shown in Tab~\ref{tab:inference_time}, our method with $5$ iterations is almost $3$ times faster than AR attention-based decoders while achieving higher recognition accuracy. When the iteration number is $1$, the inference time becomes comparable to the CTC-based decoder since they are both fully parallel decoding. In other words, our method is flexible between efficiency and accuracy. The iteration number can be adjusted based on the efficiency and accuracy requirements of different real-world applications.

\subsection{Qualitative Analysis}
\subsubsection{Easy First Decoding}

As shown in Fig.~\ref{fig:chinese_iter} We also provide a depiction of the easy-first decoding process employed by IPAD for Chinese recognition. Given the intricate nature of Chinese texts, which often operate at the sentence level without definitive word boundaries, understanding context becomes paramount. The combined prowess of the easy-first and discrete diffusion strategies enables our model to harness the inherent contextual ties between characters, refining initial, less accurate predictions. A case in point is the first image: an optical illusion initially led the model astray, resulting in the prediction ``\begin{CJK}{UTF8}{gbsn}人口防量·高人口素素\end{CJK}". However, subsequent iterations capitalized on the contextual cues, leading to improved accuracy. The third character ``\begin{CJK}{UTF8}{gbsn}数\end{CJK}" eventually gets recognized, aided by the surrounding characters.

\subsubsection{Chinese Recognition}
In light of the outstanding performance exhibited by our IPAD compared to other state-of-the-art methods, especially on the Scene subset of BCTR, we undertook qualitative comparisons against previous methodologies. Figure~\ref{fig:chinese_samples} offers illustrative examples, underscoring IPAD's robustness in recognizing occluded texts, texts in non-standard fonts, and texts embedded in intricate backgrounds. As illustrated, most previous methods recognize the final image in the first row wrongly because it is influenced by significant disturbances. They also do not perform well on the inaugural image in the second row with artistic text formations. However, these two images are interpreted by our IPAD correctly.

\section{Conclusion}
In this paper, we propose a Parallel, Iterative, and Diffusion-based network for STR, which utilizes the easy-first strategy to refine the coarse prediction iteratively and successfully balance the accuracy and efficiency of scene text recognition. To further improve the contextual learning ability of the parallel decoder, the IPAD enables the parallel decoder to learn more abundant bidirectional contextual information with the discrete diffusion strategy by viewing STR as an image-based conditional text generation process.
Extensive experiments demonstrate that our model achieves comparable results on most English benchmarks and a faster inference speed compared with other state-of-the-art autoregressive methods. Besides, it boosts the performance of BCTR datasets for Chinese recognition, especially the Scene subset. In the future, we will explore the potential of introducing the concept of large language models in NLP to our model.

\bmhead{Acknowledgements}

Supported by the National Natural Science Foundation of China (Grant NO 62376266), and by the Key Research Program of Frontier Sciences, CAS (Grant NO ZDBS-LY-7024).

\bmhead{Data Availability Statement}
The data that support the findings of this study are openly available from the corresponding reference papers.





\bibliography{sn-bibliography}


\begin{thebibliography}{113}
\ifx \bisbn   \undefined \def \bisbn  #1{ISBN #1}\fi
\ifx \binits  \undefined \def \binits#1{#1}\fi
\ifx \bauthor  \undefined \def \bauthor#1{#1}\fi
\ifx \batitle  \undefined \def \batitle#1{#1}\fi
\ifx \bjtitle  \undefined \def \bjtitle#1{#1}\fi
\ifx \bvolume  \undefined \def \bvolume#1{\textbf{#1}}\fi
\ifx \byear  \undefined \def \byear#1{#1}\fi
\ifx \bissue  \undefined \def \bissue#1{#1}\fi
\ifx \bfpage  \undefined \def \bfpage#1{#1}\fi
\ifx \blpage  \undefined \def \blpage #1{#1}\fi
\ifx \burl  \undefined \def \burl#1{\textsf{#1}}\fi
\ifx \doiurl  \undefined \def \doiurl#1{\url{https://doi.org/#1}}\fi
\ifx \betal  \undefined \def \betal{\textit{et al.}}\fi
\ifx \binstitute  \undefined \def \binstitute#1{#1}\fi
\ifx \binstitutionaled  \undefined \def \binstitutionaled#1{#1}\fi
\ifx \bctitle  \undefined \def \bctitle#1{#1}\fi
\ifx \beditor  \undefined \def \beditor#1{#1}\fi
\ifx \bpublisher  \undefined \def \bpublisher#1{#1}\fi
\ifx \bbtitle  \undefined \def \bbtitle#1{#1}\fi
\ifx \bedition  \undefined \def \bedition#1{#1}\fi
\ifx \bseriesno  \undefined \def \bseriesno#1{#1}\fi
\ifx \blocation  \undefined \def \blocation#1{#1}\fi
\ifx \bsertitle  \undefined \def \bsertitle#1{#1}\fi
\ifx \bsnm \undefined \def \bsnm#1{#1}\fi
\ifx \bsuffix \undefined \def \bsuffix#1{#1}\fi
\ifx \bparticle \undefined \def \bparticle#1{#1}\fi
\ifx \barticle \undefined \def \barticle#1{#1}\fi
\bibcommenthead
\ifx \bconfdate \undefined \def \bconfdate #1{#1}\fi
\ifx \botherref \undefined \def \botherref #1{#1}\fi
\ifx \url \undefined \def \url#1{\textsf{#1}}\fi
\ifx \bchapter \undefined \def \bchapter#1{#1}\fi
\ifx \bbook \undefined \def \bbook#1{#1}\fi
\ifx \bcomment \undefined \def \bcomment#1{#1}\fi
\ifx \oauthor \undefined \def \oauthor#1{#1}\fi
\ifx \citeauthoryear \undefined \def \citeauthoryear#1{#1}\fi
\ifx \endbibitem  \undefined \def \endbibitem {}\fi
\ifx \bconflocation  \undefined \def \bconflocation#1{#1}\fi
\ifx \arxivurl  \undefined \def \arxivurl#1{\textsf{#1}}\fi
\csname PreBibitemsHook\endcsname

\bibitem[\protect\citeauthoryear{Zhou et~al.}{2017}]{zhou2017east}
\begin{bchapter}
\bauthor{\bsnm{Zhou}, \binits{X.}},
\bauthor{\bsnm{Yao}, \binits{C.}},
\bauthor{\bsnm{Wen}, \binits{H.}},
\bauthor{\bsnm{Wang}, \binits{Y.}},
\bauthor{\bsnm{Zhou}, \binits{S.}},
\bauthor{\bsnm{He}, \binits{W.}},
\bauthor{\bsnm{Liang}, \binits{J.}}:
\bctitle{{EAST}: An efficient and accurate scene text detector}.
In: \bbtitle{Proceedings of the IEEE Conference on Computer Vision and Pattern Recognition},
pp. \bfpage{5551}--\blpage{5560}
(\byear{2017})
\end{bchapter}
\endbibitem

\bibitem[\protect\citeauthoryear{Shi et~al.}{2016}]{shi2016end}
\begin{barticle}
\bauthor{\bsnm{Shi}, \binits{B.}},
\bauthor{\bsnm{Bai}, \binits{X.}},
\bauthor{\bsnm{Yao}, \binits{C.}}:
\batitle{An end-to-end trainable neural network for image-based sequence recognition and its application to scene text recognition}.
\bjtitle{IEEE Transactions on Pattern Analysis and Machine Intelligence}
\bvolume{39}(\bissue{11}),
\bfpage{2298}--\blpage{2304}
(\byear{2016})
\end{barticle}
\endbibitem

\bibitem[\protect\citeauthoryear{Wang et~al.}{2022}]{wang2022tpsnet}
\begin{bchapter}
\bauthor{\bsnm{Wang}, \binits{W.}},
\bauthor{\bsnm{Zhou}, \binits{Y.}},
\bauthor{\bsnm{Lv}, \binits{J.}},
\bauthor{\bsnm{Wu}, \binits{D.}},
\bauthor{\bsnm{Zhao}, \binits{G.}},
\bauthor{\bsnm{Jiang}, \binits{N.}},
\bauthor{\bsnm{Wang}, \binits{W.}}:
\bctitle{{TPSNet}: Reverse thinking of thin plate splines for arbitrary shape scene text representation}.
In: \bbtitle{Proceedings of the 30th ACM International Conference on Multimedia},
pp. \bfpage{5014}--\blpage{5025}
(\byear{2022})
\end{bchapter}
\endbibitem

\bibitem[\protect\citeauthoryear{Zeng et~al.}{2023}]{zeng2023beyond}
\begin{barticle}
\bauthor{\bsnm{Zeng}, \binits{G.}},
\bauthor{\bsnm{Zhang}, \binits{Y.}},
\bauthor{\bsnm{Zhou}, \binits{Y.}},
\bauthor{\bsnm{Yang}, \binits{X.}},
\bauthor{\bsnm{Jiang}, \binits{N.}},
\bauthor{\bsnm{Zhao}, \binits{G.}},
\bauthor{\bsnm{Wang}, \binits{W.}},
\bauthor{\bsnm{Yin}, \binits{X.-C.}}:
\batitle{{Beyond OCR+VQA: T}owards end-to-end reading and reasoning for robust and accurate textvqa}.
\bjtitle{Pattern Recognition}
\bvolume{138},
\bfpage{109337}
(\byear{2023})
\end{barticle}
\endbibitem

\bibitem[\protect\citeauthoryear{He et~al.}{2016}]{he2016reading}
\begin{bchapter}
\bauthor{\bsnm{He}, \binits{P.}},
\bauthor{\bsnm{Huang}, \binits{W.}},
\bauthor{\bsnm{Qiao}, \binits{Y.}},
\bauthor{\bsnm{Loy}, \binits{C.}},
\bauthor{\bsnm{Tang}, \binits{X.}}:
\bctitle{Reading scene text in deep convolutional sequences}.
In: \bbtitle{Proceedings of the AAAI Conference on Artificial Intelligence},
vol. \bseriesno{30}
(\byear{2016})
\end{bchapter}
\endbibitem

\bibitem[\protect\citeauthoryear{Su and Lu}{2017}]{su2017accurate}
\begin{barticle}
\bauthor{\bsnm{Su}, \binits{B.}},
\bauthor{\bsnm{Lu}, \binits{S.}}:
\batitle{Accurate recognition of words in scenes without character segmentation using recurrent neural network}.
\bjtitle{Pattern Recognition}
\bvolume{63},
\bfpage{397}--\blpage{405}
(\byear{2017})
\end{barticle}
\endbibitem

\bibitem[\protect\citeauthoryear{Wang and Hu}{2017}]{wang2017gated}
\begin{botherref}
\oauthor{\bsnm{Wang}, \binits{J.}},
\oauthor{\bsnm{Hu}, \binits{X.}}:
Gated recurrent convolution neural network for ocr.
Advances in Neural Information Processing Systems
\textbf{30}
(2017)
\end{botherref}
\endbibitem

\bibitem[\protect\citeauthoryear{Du et~al.}{2022}]{du2022svtr}
\begin{bchapter}
\bauthor{\bsnm{Du}, \binits{Y.}},
\bauthor{\bsnm{Chen}, \binits{Z.}},
\bauthor{\bsnm{Jia}, \binits{C.}},
\bauthor{\bsnm{Yin}, \binits{X.}},
\bauthor{\bsnm{Zheng}, \binits{T.}},
\bauthor{\bsnm{Li}, \binits{C.}},
\bauthor{\bsnm{Du}, \binits{Y.}},
\bauthor{\bsnm{Jiang}, \binits{Y.-G.}}:
\bctitle{{SVTR: S}cene text recognition with a single visual model}.
In: \bbtitle{Proceedings of the 31st International Joint Conference on Artificial Intelligence},
pp. \bfpage{884}--\blpage{890}
(\byear{2022})
\end{bchapter}
\endbibitem

\bibitem[\protect\citeauthoryear{Shi et~al.}{2018}]{shi2018aster}
\begin{barticle}
\bauthor{\bsnm{Shi}, \binits{B.}},
\bauthor{\bsnm{Yang}, \binits{M.}},
\bauthor{\bsnm{Wang}, \binits{X.}},
\bauthor{\bsnm{Lyu}, \binits{P.}},
\bauthor{\bsnm{Yao}, \binits{C.}},
\bauthor{\bsnm{Bai}, \binits{X.}}:
\batitle{{ASTER: An} attentional scene text recognizer with flexible rectification}.
\bjtitle{IEEE Transactions on Pattern Analysis and Machine Intelligence}
\bvolume{41}(\bissue{9}),
\bfpage{2035}--\blpage{2048}
(\byear{2018})
\end{barticle}
\endbibitem

\bibitem[\protect\citeauthoryear{Baek et~al.}{2019}]{baek2019wrong}
\begin{bchapter}
\bauthor{\bsnm{Baek}, \binits{J.}},
\bauthor{\bsnm{Kim}, \binits{G.}},
\bauthor{\bsnm{Lee}, \binits{J.}},
\bauthor{\bsnm{Park}, \binits{S.}},
\bauthor{\bsnm{Han}, \binits{D.}},
\bauthor{\bsnm{Yun}, \binits{S.}},
\bauthor{\bsnm{Oh}, \binits{S.J.}},
\bauthor{\bsnm{Lee}, \binits{H.}}:
\bctitle{What is wrong with scene text recognition model comparisons? {Dataset} and model analysis}.
In: \bbtitle{Proceedings of the IEEE/CVF International Conference on Computer Vision},
pp. \bfpage{4715}--\blpage{4723}
(\byear{2019})
\end{bchapter}
\endbibitem

\bibitem[\protect\citeauthoryear{Li et~al.}{2019}]{li2019show}
\begin{bchapter}
\bauthor{\bsnm{Li}, \binits{H.}},
\bauthor{\bsnm{Wang}, \binits{P.}},
\bauthor{\bsnm{Shen}, \binits{C.}},
\bauthor{\bsnm{Zhang}, \binits{G.}}:
\bctitle{Show, attend and read: A simple and strong baseline for irregular text recognition}.
In: \bbtitle{Proceedings of the AAAI Conference on Artificial Intelligence},
pp. \bfpage{8610}--\blpage{8617}
(\byear{2019})
\end{bchapter}
\endbibitem

\bibitem[\protect\citeauthoryear{Yu et~al.}{2020}]{yu2020towards}
\begin{bchapter}
\bauthor{\bsnm{Yu}, \binits{D.}},
\bauthor{\bsnm{Li}, \binits{X.}},
\bauthor{\bsnm{Zhang}, \binits{C.}},
\bauthor{\bsnm{Liu}, \binits{T.}},
\bauthor{\bsnm{Han}, \binits{J.}},
\bauthor{\bsnm{Liu}, \binits{J.}},
\bauthor{\bsnm{Ding}, \binits{E.}}:
\bctitle{Towards accurate scene text recognition with semantic reasoning networks}.
In: \bbtitle{Proceedings of the IEEE/CVF Conference on Computer Vision and Pattern Recognition},
pp. \bfpage{12113}--\blpage{12122}
(\byear{2020})
\end{bchapter}
\endbibitem

\bibitem[\protect\citeauthoryear{Bautista and Atienza}{2022}]{bautista2022scene}
\begin{bchapter}
\bauthor{\bsnm{Bautista}, \binits{D.}},
\bauthor{\bsnm{Atienza}, \binits{R.}}:
\bctitle{Scene text recognition with permuted autoregressive sequence models}.
In: \bbtitle{Proceedings of the European Conference on Computer Vision},
pp. \bfpage{178}--\blpage{196}
(\byear{2022})
\end{bchapter}
\endbibitem

\bibitem[\protect\citeauthoryear{Liao et~al.}{2019}]{liao2019scene}
\begin{bchapter}
\bauthor{\bsnm{Liao}, \binits{M.}},
\bauthor{\bsnm{Zhang}, \binits{J.}},
\bauthor{\bsnm{Wan}, \binits{Z.}},
\bauthor{\bsnm{Xie}, \binits{F.}},
\bauthor{\bsnm{Liang}, \binits{J.}},
\bauthor{\bsnm{Lyu}, \binits{P.}},
\bauthor{\bsnm{Yao}, \binits{C.}},
\bauthor{\bsnm{Bai}, \binits{X.}}:
\bctitle{Scene text recognition from two-dimensional perspective}.
In: \bbtitle{Proceedings of the AAAI Conference on Artificial Intelligence},
vol. \bseriesno{33},
pp. \bfpage{8714}--\blpage{8721}
(\byear{2019})
\end{bchapter}
\endbibitem

\bibitem[\protect\citeauthoryear{Wan et~al.}{2020}]{wan2020textscanner}
\begin{bchapter}
\bauthor{\bsnm{Wan}, \binits{Z.}},
\bauthor{\bsnm{He}, \binits{M.}},
\bauthor{\bsnm{Chen}, \binits{H.}},
\bauthor{\bsnm{Bai}, \binits{X.}},
\bauthor{\bsnm{Yao}, \binits{C.}}:
\bctitle{Textscanner: Reading characters in order for robust scene text recognition}.
In: \bbtitle{Proceedings of the AAAI Conference on Artificial Intelligence},
vol. \bseriesno{34},
pp. \bfpage{12120}--\blpage{12127}
(\byear{2020})
\end{bchapter}
\endbibitem

\bibitem[\protect\citeauthoryear{Zhong et~al.}{2022}]{zhong2022sgbanet}
\begin{bchapter}
\bauthor{\bsnm{Zhong}, \binits{D.}},
\bauthor{\bsnm{Lyu}, \binits{S.}},
\bauthor{\bsnm{Shivakumara}, \binits{P.}},
\bauthor{\bsnm{Yin}, \binits{B.}},
\bauthor{\bsnm{Wu}, \binits{J.}},
\bauthor{\bsnm{Pal}, \binits{U.}},
\bauthor{\bsnm{Lu}, \binits{Y.}}:
\bctitle{{SGBANet: Semantic GAN} and balanced attention network for arbitrarily oriented scene text recognition}.
In: \bbtitle{Proceedings of the European Conference on Computer Vision},
pp. \bfpage{464}--\blpage{480}
(\byear{2022})
\end{bchapter}
\endbibitem

\bibitem[\protect\citeauthoryear{Yang et~al.}{2024}]{yang2024masked}
\begin{botherref}
\oauthor{\bsnm{Yang}, \binits{X.}},
\oauthor{\bsnm{Qiao}, \binits{Z.}},
\oauthor{\bsnm{Wei}, \binits{J.}},
\oauthor{\bsnm{Yang}, \binits{D.}},
\oauthor{\bsnm{Zhou}, \binits{Y.}}:
Masked and permuted implicit context learning for scene text recognition.
IEEE Signal Processing Letters
(2024)
\end{botherref}
\endbibitem

\bibitem[\protect\citeauthoryear{Fang et~al.}{2021}]{fang2021read}
\begin{bchapter}
\bauthor{\bsnm{Fang}, \binits{S.}},
\bauthor{\bsnm{Xie}, \binits{H.}},
\bauthor{\bsnm{Wang}, \binits{Y.}},
\bauthor{\bsnm{Mao}, \binits{Z.}},
\bauthor{\bsnm{Zhang}, \binits{Y.}}:
\bctitle{Read like humans: Autonomous, bidirectional and iterative language modeling for scene text recognition}.
In: \bbtitle{Proceedings of the IEEE/CVF Conference on Computer Vision and Pattern Recognition},
pp. \bfpage{7098}--\blpage{7107}
(\byear{2021})
\end{bchapter}
\endbibitem

\bibitem[\protect\citeauthoryear{Goldberg and Elhadad}{2010}]{goldberg2010efficient}
\begin{bchapter}
\bauthor{\bsnm{Goldberg}, \binits{Y.}},
\bauthor{\bsnm{Elhadad}, \binits{M.}}:
\bctitle{An efficient algorithm for easy-first non-directional dependency parsing}.
In: \bbtitle{Human Language Technologies: The 2010 Annual Conference of the North American Chapter of the Association for Computational Linguistics},
pp. \bfpage{742}--\blpage{750}
(\byear{2010})
\end{bchapter}
\endbibitem

\bibitem[\protect\citeauthoryear{Vaswani et~al.}{2017}]{vaswani2017attention}
\begin{botherref}
\oauthor{\bsnm{Vaswani}, \binits{A.}},
\oauthor{\bsnm{Shazeer}, \binits{N.}},
\oauthor{\bsnm{Parmar}, \binits{N.}},
\oauthor{\bsnm{Uszkoreit}, \binits{J.}},
\oauthor{\bsnm{Jones}, \binits{L.}},
\oauthor{\bsnm{Gomez}, \binits{A.N.}},
\oauthor{\bsnm{Kaiser}, \binits{{\L}.}},
\oauthor{\bsnm{Polosukhin}, \binits{I.}}:
Attention is all you need.
Advances in Neural Information Processing Systems
\textbf{30}
(2017)
\end{botherref}
\endbibitem

\bibitem[\protect\citeauthoryear{Dhariwal and Nichol}{2021}]{dhariwal2021diffusion}
\begin{barticle}
\bauthor{\bsnm{Dhariwal}, \binits{P.}},
\bauthor{\bsnm{Nichol}, \binits{A.}}:
\batitle{Diffusion models beat {GANs} on image synthesis}.
\bjtitle{Advances in Neural Information Processing Systems}
\bvolume{34},
\bfpage{8780}--\blpage{8794}
(\byear{2021})
\end{barticle}
\endbibitem

\bibitem[\protect\citeauthoryear{Rombach et~al.}{2022}]{rombach2022high}
\begin{bchapter}
\bauthor{\bsnm{Rombach}, \binits{R.}},
\bauthor{\bsnm{Blattmann}, \binits{A.}},
\bauthor{\bsnm{Lorenz}, \binits{D.}},
\bauthor{\bsnm{Esser}, \binits{P.}},
\bauthor{\bsnm{Ommer}, \binits{B.}}:
\bctitle{High-resolution image synthesis with latent diffusion models}.
In: \bbtitle{Proceedings of the IEEE/CVF Conference on Computer Vision and Pattern Recognition},
pp. \bfpage{10684}--\blpage{10695}
(\byear{2022})
\end{bchapter}
\endbibitem

\bibitem[\protect\citeauthoryear{He et~al.}{2023}]{he2022diffusionbert}
\begin{bchapter}
\bauthor{\bsnm{He}, \binits{Z.}},
\bauthor{\bsnm{Sun}, \binits{T.}},
\bauthor{\bsnm{Tang}, \binits{Q.}},
\bauthor{\bsnm{Wang}, \binits{K.}},
\bauthor{\bsnm{Huang}, \binits{X.}},
\bauthor{\bsnm{Qiu}, \binits{X.}}:
\bctitle{{D}iffusion{BERT}: Improving generative masked language models with diffusion models}.
In: \bbtitle{Proceedings of the 61st Annual Meeting of the Association for Computational Linguistics},
pp. \bfpage{4521}--\blpage{4534}
(\byear{2023})
\end{bchapter}
\endbibitem

\bibitem[\protect\citeauthoryear{Qiao et~al.}{2021}]{qiao2021pimnet}
\begin{bchapter}
\bauthor{\bsnm{Qiao}, \binits{Z.}},
\bauthor{\bsnm{Zhou}, \binits{Y.}},
\bauthor{\bsnm{Wei}, \binits{J.}},
\bauthor{\bsnm{Wang}, \binits{W.}},
\bauthor{\bsnm{Zhang}, \binits{Y.}},
\bauthor{\bsnm{Jiang}, \binits{N.}},
\bauthor{\bsnm{Wang}, \binits{H.}},
\bauthor{\bsnm{Wang}, \binits{W.}}:
\bctitle{{PIMNet}: A parallel, iterative and mimicking network for scene text recognition}.
In: \bbtitle{Proceedings of the 29th ACM International Conference on Multimedia},
pp. \bfpage{2046}--\blpage{2055}
(\byear{2021})
\end{bchapter}
\endbibitem

\bibitem[\protect\citeauthoryear{Mishra et~al.}{2012a}]{mishra2012scene}
\begin{bchapter}
\bauthor{\bsnm{Mishra}, \binits{A.}},
\bauthor{\bsnm{Alahari}, \binits{K.}},
\bauthor{\bsnm{Jawahar}, \binits{C.}}:
\bctitle{Scene text recognition using higher order language priors}.
In: \bbtitle{British Machine Vision Conference}
(\byear{2012})
\end{bchapter}
\endbibitem

\bibitem[\protect\citeauthoryear{Mishra et~al.}{2012b}]{mishra2012top}
\begin{bchapter}
\bauthor{\bsnm{Mishra}, \binits{A.}},
\bauthor{\bsnm{Alahari}, \binits{K.}},
\bauthor{\bsnm{Jawahar}, \binits{C.}}:
\bctitle{Top-down and bottom-up cues for scene text recognition}.
In: \bbtitle{2012 IEEE Conference on Computer Vision and Pattern Recognition},
pp. \bfpage{2687}--\blpage{2694}
(\byear{2012}).
\bcomment{IEEE}
\end{bchapter}
\endbibitem

\bibitem[\protect\citeauthoryear{Neumann and Matas}{2012}]{neumann2012real}
\begin{bchapter}
\bauthor{\bsnm{Neumann}, \binits{L.}},
\bauthor{\bsnm{Matas}, \binits{J.}}:
\bctitle{Real-time scene text localization and recognition}.
In: \bbtitle{Proceedings of the IEEE/CVF Conference on Computer Vision and Pattern Recognition},
pp. \bfpage{3538}--\blpage{3545}
(\byear{2012})
\end{bchapter}
\endbibitem

\bibitem[\protect\citeauthoryear{Novikova et~al.}{2012}]{novikova2012large}
\begin{bchapter}
\bauthor{\bsnm{Novikova}, \binits{T.}},
\bauthor{\bsnm{Barinova}, \binits{O.}},
\bauthor{\bsnm{Kohli}, \binits{P.}},
\bauthor{\bsnm{Lempitsky}, \binits{V.}}:
\bctitle{Large-lexicon attribute-consistent text recognition in natural images}.
In: \bbtitle{Computer Vision--ECCV 2012: 12th European Conference on Computer Vision, Florence, Italy, October 7-13, 2012, Proceedings, Part VI 12},
pp. \bfpage{752}--\blpage{765}
(\byear{2012}).
\bcomment{Springer}
\end{bchapter}
\endbibitem

\bibitem[\protect\citeauthoryear{Wang et~al.}{2011}]{wang2011end}
\begin{bchapter}
\bauthor{\bsnm{Wang}, \binits{K.}},
\bauthor{\bsnm{Babenko}, \binits{B.}},
\bauthor{\bsnm{Belongie}, \binits{S.}}:
\bctitle{End-to-end scene text recognition}.
In: \bbtitle{Proceedings of the IEEE International Conference on Computer Vision},
pp. \bfpage{1457}--\blpage{1464}
(\byear{2011})
\end{bchapter}
\endbibitem

\bibitem[\protect\citeauthoryear{Wang and Belongie}{2010}]{wang2010word}
\begin{bchapter}
\bauthor{\bsnm{Wang}, \binits{K.}},
\bauthor{\bsnm{Belongie}, \binits{S.}}:
\bctitle{Word spotting in the wild}.
In: \bbtitle{Computer Vision--ECCV 2010: 11th European Conference on Computer Vision, Heraklion, Crete, Greece, September 5-11, 2010, Proceedings, Part I 11},
pp. \bfpage{591}--\blpage{604}
(\byear{2010}).
\bcomment{Springer}
\end{bchapter}
\endbibitem

\bibitem[\protect\citeauthoryear{Wang et~al.}{2012}]{wang2012end}
\begin{bchapter}
\bauthor{\bsnm{Wang}, \binits{T.}},
\bauthor{\bsnm{Wu}, \binits{D.J.}},
\bauthor{\bsnm{Coates}, \binits{A.}},
\bauthor{\bsnm{Ng}, \binits{A.Y.}}:
\bctitle{End-to-end text recognition with convolutional neural networks}.
In: \bbtitle{Proceedings of the 21st International Conference on Pattern Recognition (ICPR2012)},
pp. \bfpage{3304}--\blpage{3308}
(\byear{2012}).
\bcomment{IEEE}
\end{bchapter}
\endbibitem

\bibitem[\protect\citeauthoryear{Yao et~al.}{2014}]{yao2014strokelets}
\begin{bchapter}
\bauthor{\bsnm{Yao}, \binits{C.}},
\bauthor{\bsnm{Bai}, \binits{X.}},
\bauthor{\bsnm{Shi}, \binits{B.}},
\bauthor{\bsnm{Liu}, \binits{W.}}:
\bctitle{Strokelets: A learned multi-scale representation for scene text recognition}.
In: \bbtitle{Proceedings of the IEEE/CVF Conference on Computer Vision and Pattern Recognition},
pp. \bfpage{4042}--\blpage{4049}
(\byear{2014})
\end{bchapter}
\endbibitem

\bibitem[\protect\citeauthoryear{Lee and Osindero}{2016}]{lee2016recursive}
\begin{bchapter}
\bauthor{\bsnm{Lee}, \binits{C.-Y.}},
\bauthor{\bsnm{Osindero}, \binits{S.}}:
\bctitle{Recursive recurrent nets with attention modeling for {OCR} in the wild}.
In: \bbtitle{Proceedings of the IEEE/CVF Conference on Computer Vision and Pattern Recognition},
pp. \bfpage{2231}--\blpage{2239}
(\byear{2016})
\end{bchapter}
\endbibitem

\bibitem[\protect\citeauthoryear{Cheng et~al.}{2017}]{cheng2017focusing}
\begin{bchapter}
\bauthor{\bsnm{Cheng}, \binits{Z.}},
\bauthor{\bsnm{Bai}, \binits{F.}},
\bauthor{\bsnm{Xu}, \binits{Y.}},
\bauthor{\bsnm{Zheng}, \binits{G.}},
\bauthor{\bsnm{Pu}, \binits{S.}},
\bauthor{\bsnm{Zhou}, \binits{S.}}:
\bctitle{Focusing attention: Towards accurate text recognition in natural images}.
In: \bbtitle{Proceedings of the IEEE/CVF International Conference on Computer Vision},
pp. \bfpage{5076}--\blpage{5084}
(\byear{2017})
\end{bchapter}
\endbibitem

\bibitem[\protect\citeauthoryear{Fang et~al.}{2018}]{fang2018attention}
\begin{bchapter}
\bauthor{\bsnm{Fang}, \binits{S.}},
\bauthor{\bsnm{Xie}, \binits{H.}},
\bauthor{\bsnm{Zha}, \binits{Z.-J.}},
\bauthor{\bsnm{Sun}, \binits{N.}},
\bauthor{\bsnm{Tan}, \binits{J.}},
\bauthor{\bsnm{Zhang}, \binits{Y.}}:
\bctitle{Attention and language ensemble for scene text recognition with convolutional sequence modeling}.
In: \bbtitle{Proceedings of the 26th ACM International Conference on Multimedia},
pp. \bfpage{248}--\blpage{256}
(\byear{2018})
\end{bchapter}
\endbibitem

\bibitem[\protect\citeauthoryear{Luo et~al.}{2019}]{luo2019moran}
\begin{barticle}
\bauthor{\bsnm{Luo}, \binits{C.}},
\bauthor{\bsnm{Jin}, \binits{L.}},
\bauthor{\bsnm{Sun}, \binits{Z.}}:
\batitle{{MORAN: A multi-object rectified attention network for scene text recognition}}.
\bjtitle{Pattern Recognition}
\bvolume{90},
\bfpage{109}--\blpage{118}
(\byear{2019})
\end{barticle}
\endbibitem

\bibitem[\protect\citeauthoryear{Shi et~al.}{2016}]{shi2016robust}
\begin{bchapter}
\bauthor{\bsnm{Shi}, \binits{B.}},
\bauthor{\bsnm{Wang}, \binits{X.}},
\bauthor{\bsnm{Lyu}, \binits{P.}},
\bauthor{\bsnm{Yao}, \binits{C.}},
\bauthor{\bsnm{Bai}, \binits{X.}}:
\bctitle{Robust scene text recognition with automatic rectification}.
In: \bbtitle{Proceedings of the IEEE/CVF Conference on Computer Vision and Pattern Recognition},
pp. \bfpage{4168}--\blpage{4176}
(\byear{2016})
\end{bchapter}
\endbibitem

\bibitem[\protect\citeauthoryear{Zhan and Lu}{2019}]{zhan2019esir}
\begin{bchapter}
\bauthor{\bsnm{Zhan}, \binits{F.}},
\bauthor{\bsnm{Lu}, \binits{S.}}:
\bctitle{{ESIR: End}-to-end scene text recognition via iterative image rectification}.
In: \bbtitle{Proceedings of the IEEE/CVF Conference on Computer Vision and Pattern Recognition},
pp. \bfpage{2059}--\blpage{2068}
(\byear{2019})
\end{bchapter}
\endbibitem

\bibitem[\protect\citeauthoryear{Yang et~al.}{2019}]{yang2019symmetry}
\begin{bchapter}
\bauthor{\bsnm{Yang}, \binits{M.}},
\bauthor{\bsnm{Guan}, \binits{Y.}},
\bauthor{\bsnm{Liao}, \binits{M.}},
\bauthor{\bsnm{He}, \binits{X.}},
\bauthor{\bsnm{Bian}, \binits{K.}},
\bauthor{\bsnm{Bai}, \binits{S.}},
\bauthor{\bsnm{Yao}, \binits{C.}},
\bauthor{\bsnm{Bai}, \binits{X.}}:
\bctitle{Symmetry-constrained rectification network for scene text recognition}.
In: \bbtitle{Proceedings of the IEEE/CVF International Conference on Computer Vision},
pp. \bfpage{9147}--\blpage{9156}
(\byear{2019})
\end{bchapter}
\endbibitem

\bibitem[\protect\citeauthoryear{Yang et~al.}{2017}]{yang2017learning}
\begin{bchapter}
\bauthor{\bsnm{Yang}, \binits{X.}},
\bauthor{\bsnm{He}, \binits{D.}},
\bauthor{\bsnm{Zhou}, \binits{Z.}},
\bauthor{\bsnm{Kifer}, \binits{D.}},
\bauthor{\bsnm{Giles}, \binits{C.L.}}:
\bctitle{Learning to read irregular text with attention mechanisms.}
In: \bbtitle{Proceedings of the 27th International Joint Conferences on Artificial Intelligence},
vol. \bseriesno{1},
p. \bfpage{3}
(\byear{2017})
\end{bchapter}
\endbibitem

\bibitem[\protect\citeauthoryear{Wang et~al.}{2020}]{wang2020decoupled}
\begin{bchapter}
\bauthor{\bsnm{Wang}, \binits{T.}},
\bauthor{\bsnm{Zhu}, \binits{Y.}},
\bauthor{\bsnm{Jin}, \binits{L.}},
\bauthor{\bsnm{Luo}, \binits{C.}},
\bauthor{\bsnm{Chen}, \binits{X.}},
\bauthor{\bsnm{Wu}, \binits{Y.}},
\bauthor{\bsnm{Wang}, \binits{Q.}},
\bauthor{\bsnm{Cai}, \binits{M.}}:
\bctitle{Decoupled attention network for text recognition}.
In: \bbtitle{Proceedings of the AAAI Conference on Artificial Intelligence},
pp. \bfpage{12216}--\blpage{12224}
(\byear{2020})
\end{bchapter}
\endbibitem

\bibitem[\protect\citeauthoryear{Qiao et~al.}{2021}]{qiao2021gaussian}
\begin{bchapter}
\bauthor{\bsnm{Qiao}, \binits{Z.}},
\bauthor{\bsnm{Qin}, \binits{X.}},
\bauthor{\bsnm{Zhou}, \binits{Y.}},
\bauthor{\bsnm{Yang}, \binits{F.}},
\bauthor{\bsnm{Wang}, \binits{W.}}:
\bctitle{Gaussian constrained attention network for scene text recognition}.
In: \bbtitle{2020 25th International Conference on Pattern Recognition (ICPR)},
pp. \bfpage{3328}--\blpage{3335}
(\byear{2021}).
\bcomment{IEEE}
\end{bchapter}
\endbibitem

\bibitem[\protect\citeauthoryear{Qiao et~al.}{2020}]{qiao2020seed}
\begin{bchapter}
\bauthor{\bsnm{Qiao}, \binits{Z.}},
\bauthor{\bsnm{Zhou}, \binits{Y.}},
\bauthor{\bsnm{Yang}, \binits{D.}},
\bauthor{\bsnm{Zhou}, \binits{Y.}},
\bauthor{\bsnm{Wang}, \binits{W.}}:
\bctitle{{SEED: S}emantics enhanced encoder-decoder framework for scene text recognition}.
In: \bbtitle{Proceedings of the IEEE/CVF Conference on Computer Vision and Pattern Recognition},
pp. \bfpage{13528}--\blpage{13537}
(\byear{2020})
\end{bchapter}
\endbibitem

\bibitem[\protect\citeauthoryear{Litman et~al.}{2020}]{litman2020scatter}
\begin{bchapter}
\bauthor{\bsnm{Litman}, \binits{R.}},
\bauthor{\bsnm{Anschel}, \binits{O.}},
\bauthor{\bsnm{Tsiper}, \binits{S.}},
\bauthor{\bsnm{Litman}, \binits{R.}},
\bauthor{\bsnm{Mazor}, \binits{S.}},
\bauthor{\bsnm{Manmatha}, \binits{R.}}:
\bctitle{{SCATTER: Selective} context attentional scene text recognizer}.
In: \bbtitle{Proceedings of the IEEE/CVF Conference on Computer Vision and Pattern Recognition},
pp. \bfpage{11962}--\blpage{11972}
(\byear{2020})
\end{bchapter}
\endbibitem

\bibitem[\protect\citeauthoryear{Zheng et~al.}{2020}]{zheng2020lal}
\begin{bchapter}
\bauthor{\bsnm{Zheng}, \binits{Y.}},
\bauthor{\bsnm{Qin}, \binits{W.}},
\bauthor{\bsnm{Wijaya}, \binits{D.}},
\bauthor{\bsnm{Betke}, \binits{M.}}:
\bctitle{Lal: Linguistically aware learning for scene text recognition}.
In: \bbtitle{Proceedings of the 28th ACM International Conference on Multimedia},
pp. \bfpage{4051}--\blpage{4059}
(\byear{2020})
\end{bchapter}
\endbibitem

\bibitem[\protect\citeauthoryear{Tan et~al.}{2022}]{tan2022pure}
\begin{bchapter}
\bauthor{\bsnm{Tan}, \binits{Y.L.}},
\bauthor{\bsnm{Kong}, \binits{A.W.-K.}},
\bauthor{\bsnm{Kim}, \binits{J.-J.}}:
\bctitle{Pure transformer with integrated experts for scene text recognition}.
In: \bbtitle{European Conference on Computer Vision},
pp. \bfpage{481}--\blpage{497}
(\byear{2022})
\end{bchapter}
\endbibitem

\bibitem[\protect\citeauthoryear{Jaderberg et~al.}{2016}]{jaderberg2016reading}
\begin{barticle}
\bauthor{\bsnm{Jaderberg}, \binits{M.}},
\bauthor{\bsnm{Simonyan}, \binits{K.}},
\bauthor{\bsnm{Vedaldi}, \binits{A.}},
\bauthor{\bsnm{Zisserman}, \binits{A.}}:
\batitle{Reading text in the wild with convolutional neural networks}.
\bjtitle{International Journal of Computer Vision}
\bvolume{116}(\bissue{1}),
\bfpage{1}--\blpage{20}
(\byear{2016})
\end{barticle}
\endbibitem

\bibitem[\protect\citeauthoryear{Atienza}{2021}]{atienza2021vision}
\begin{bchapter}
\bauthor{\bsnm{Atienza}, \binits{R.}}:
\bctitle{Vision transformer for fast and efficient scene text recognition}.
In: \bbtitle{International Conference on Document Analysis and Recognition},
pp. \bfpage{319}--\blpage{334}
(\byear{2021})
\end{bchapter}
\endbibitem

\bibitem[\protect\citeauthoryear{Dosovitskiy et~al.}{2020}]{dosovitskiy2020image}
\begin{bchapter}
\bauthor{\bsnm{Dosovitskiy}, \binits{A.}},
\bauthor{\bsnm{Beyer}, \binits{L.}},
\bauthor{\bsnm{Kolesnikov}, \binits{A.}},
\bauthor{\bsnm{Weissenborn}, \binits{D.}},
\bauthor{\bsnm{Zhai}, \binits{X.}},
\bauthor{\bsnm{Unterthiner}, \binits{T.}},
\bauthor{\bsnm{Dehghani}, \binits{M.}},
\bauthor{\bsnm{Minderer}, \binits{M.}},
\bauthor{\bsnm{Heigold}, \binits{G.}},
\bauthor{\bsnm{Gelly}, \binits{S.}}, \betal:
\bctitle{An image is worth 16x16 words: Transformers for image recognition at scale}.
In: \bbtitle{International Conference on Learning Representations}
(\byear{2020})
\end{bchapter}
\endbibitem

\bibitem[\protect\citeauthoryear{Chao et~al.}{2020}]{chao2020variational}
\begin{bchapter}
\bauthor{\bsnm{Chao}, \binits{L.}},
\bauthor{\bsnm{Chen}, \binits{J.}},
\bauthor{\bsnm{Chu}, \binits{W.}}:
\bctitle{Variational connectionist temporal classification}.
In: \bbtitle{Proceedings of the European Conference on Computer Vision},
pp. \bfpage{460}--\blpage{476}
(\byear{2020})
\end{bchapter}
\endbibitem

\bibitem[\protect\citeauthoryear{Feng et~al.}{2019}]{feng2019textdragon}
\begin{bchapter}
\bauthor{\bsnm{Feng}, \binits{W.}},
\bauthor{\bsnm{He}, \binits{W.}},
\bauthor{\bsnm{Yin}, \binits{F.}},
\bauthor{\bsnm{Zhang}, \binits{X.-Y.}},
\bauthor{\bsnm{Liu}, \binits{C.-L.}}:
\bctitle{Textdragon: An end-to-end framework for arbitrary shaped text spotting}.
In: \bbtitle{Proceedings of the IEEE/CVF International Conference on Computer Vision},
pp. \bfpage{9076}--\blpage{9085}
(\byear{2019})
\end{bchapter}
\endbibitem

\bibitem[\protect\citeauthoryear{Hu et~al.}{2020}]{hu2020gtc}
\begin{bchapter}
\bauthor{\bsnm{Hu}, \binits{W.}},
\bauthor{\bsnm{Cai}, \binits{X.}},
\bauthor{\bsnm{Hou}, \binits{J.}},
\bauthor{\bsnm{Yi}, \binits{S.}},
\bauthor{\bsnm{Lin}, \binits{Z.}}:
\bctitle{{GTC: Guided training of CTC towards efficient and accurate scene text recognition}}.
In: \bbtitle{Proceedings of the AAAI Conference on Artificial Intelligence},
vol. \bseriesno{34},
pp. \bfpage{11005}--\blpage{11012}
(\byear{2020})
\end{bchapter}
\endbibitem

\bibitem[\protect\citeauthoryear{Guan et~al.}{2022}]{guan2022glyph}
\begin{botherref}
\oauthor{\bsnm{Guan}, \binits{T.}},
\oauthor{\bsnm{Gu}, \binits{C.}},
\oauthor{\bsnm{Tu}, \binits{J.}},
\oauthor{\bsnm{Yang}, \binits{X.}},
\oauthor{\bsnm{Feng}, \binits{Q.}}:
A glyph-driven topology enhancement network for scene text recognition.
arXiv preprint arXiv:2203.03382
(2022)
\end{botherref}
\endbibitem

\bibitem[\protect\citeauthoryear{Ghazvininejad et~al.}{2019}]{ghazvininejad2019mask}
\begin{botherref}
\oauthor{\bsnm{Ghazvininejad}, \binits{M.}},
\oauthor{\bsnm{Levy}, \binits{O.}},
\oauthor{\bsnm{Liu}, \binits{Y.}},
\oauthor{\bsnm{Zettlemoyer}, \binits{L.}}:
Mask-predict: Parallel decoding of conditional masked language models.
arXiv preprint arXiv:1904.09324
(2019)
\end{botherref}
\endbibitem

\bibitem[\protect\citeauthoryear{Gu et~al.}{2017}]{gu2017non}
\begin{botherref}
\oauthor{\bsnm{Gu}, \binits{J.}},
\oauthor{\bsnm{Bradbury}, \binits{J.}},
\oauthor{\bsnm{Xiong}, \binits{C.}},
\oauthor{\bsnm{Li}, \binits{V.O.}},
\oauthor{\bsnm{Socher}, \binits{R.}}:
Non-autoregressive neural machine translation.
arXiv preprint arXiv:1711.02281
(2017)
\end{botherref}
\endbibitem

\bibitem[\protect\citeauthoryear{Qian et~al.}{2021}]{qian-etal-2021-glancing}
\begin{bchapter}
\bauthor{\bsnm{Qian}, \binits{L.}},
\bauthor{\bsnm{Zhou}, \binits{H.}},
\bauthor{\bsnm{Bao}, \binits{Y.}},
\bauthor{\bsnm{Wang}, \binits{M.}},
\bauthor{\bsnm{Qiu}, \binits{L.}},
\bauthor{\bsnm{Zhang}, \binits{W.}},
\bauthor{\bsnm{Yu}, \binits{Y.}},
\bauthor{\bsnm{Li}, \binits{L.}}:
\bctitle{Glancing transformer for non-autoregressive neural machine translation}.
In: \bbtitle{Proceedings of the 59th Annual Meeting of the Association for Computational Linguistics},
pp. \bfpage{1993}--\blpage{2003}
(\byear{2021})
\end{bchapter}
\endbibitem

\bibitem[\protect\citeauthoryear{Chan et~al.}{2020}]{chan2020imputer}
\begin{bchapter}
\bauthor{\bsnm{Chan}, \binits{W.}},
\bauthor{\bsnm{Saharia}, \binits{C.}},
\bauthor{\bsnm{Hinton}, \binits{G.}},
\bauthor{\bsnm{Norouzi}, \binits{M.}},
\bauthor{\bsnm{Jaitly}, \binits{N.}}:
\bctitle{Imputer: Sequence modelling via imputation and dynamic programming}.
In: \bbtitle{International Conference on Machine Learning},
pp. \bfpage{1403}--\blpage{1413}
(\byear{2020}).
\bcomment{PMLR}
\end{bchapter}
\endbibitem

\bibitem[\protect\citeauthoryear{Tian et~al.}{2020}]{tian2020spike}
\begin{botherref}
\oauthor{\bsnm{Tian}, \binits{Z.}},
\oauthor{\bsnm{Yi}, \binits{J.}},
\oauthor{\bsnm{Tao}, \binits{J.}},
\oauthor{\bsnm{Bai}, \binits{Y.}},
\oauthor{\bsnm{Zhang}, \binits{S.}},
\oauthor{\bsnm{Wen}, \binits{Z.}}:
Spike-triggered non-autoregressive transformer for end-to-end speech recognition.
arXiv preprint arXiv:2005.07903
(2020)
\end{botherref}
\endbibitem

\bibitem[\protect\citeauthoryear{Chi et~al.}{2021}]{chi-etal-2021-align}
\begin{bchapter}
\bauthor{\bsnm{Chi}, \binits{E.A.}},
\bauthor{\bsnm{Salazar}, \binits{J.}},
\bauthor{\bsnm{Kirchhoff}, \binits{K.}}:
\bctitle{Align-refine: Non-autoregressive speech recognition via iterative realignment}.
In: \bbtitle{Proceedings of the 2021 Conference of the North American Chapter of the Association for Computational Linguistics: Human Language Technologies},
pp. \bfpage{1920}--\blpage{1927}
(\byear{2021})
\end{bchapter}
\endbibitem

\bibitem[\protect\citeauthoryear{Guo et~al.}{2020}]{guo2020non}
\begin{botherref}
\oauthor{\bsnm{Guo}, \binits{L.}},
\oauthor{\bsnm{Liu}, \binits{J.}},
\oauthor{\bsnm{Zhu}, \binits{X.}},
\oauthor{\bsnm{He}, \binits{X.}},
\oauthor{\bsnm{Jiang}, \binits{J.}},
\oauthor{\bsnm{Lu}, \binits{H.}}:
Non-autoregressive image captioning with counterfactuals-critical multi-agent learning.
arXiv preprint arXiv:2005.04690
(2020)
\end{botherref}
\endbibitem

\bibitem[\protect\citeauthoryear{Da et~al.}{2022}]{da2022levenshtein}
\begin{bchapter}
\bauthor{\bsnm{Da}, \binits{C.}},
\bauthor{\bsnm{Wang}, \binits{P.}},
\bauthor{\bsnm{Yao}, \binits{C.}}:
\bctitle{{Levenshtein OCR}}.
In: \bbtitle{Proceedings of the European Conference on Computer Vision},
pp. \bfpage{322}--\blpage{338}
(\byear{2022})
\end{bchapter}
\endbibitem

\bibitem[\protect\citeauthoryear{Wang et~al.}{2021}]{wang2021two}
\begin{bchapter}
\bauthor{\bsnm{Wang}, \binits{Y.}},
\bauthor{\bsnm{Xie}, \binits{H.}},
\bauthor{\bsnm{Fang}, \binits{S.}},
\bauthor{\bsnm{Wang}, \binits{J.}},
\bauthor{\bsnm{Zhu}, \binits{S.}},
\bauthor{\bsnm{Zhang}, \binits{Y.}}:
\bctitle{From two to one: A new scene text recognizer with visual language modeling network}.
In: \bbtitle{Proceedings of the IEEE/CVF International Conference on Computer Vision},
pp. \bfpage{14194}--\blpage{14203}
(\byear{2021})
\end{bchapter}
\endbibitem

\bibitem[\protect\citeauthoryear{Wang et~al.}{2022}]{wang2022multi}
\begin{bchapter}
\bauthor{\bsnm{Wang}, \binits{P.}},
\bauthor{\bsnm{Da}, \binits{C.}},
\bauthor{\bsnm{Yao}, \binits{C.}}:
\bctitle{Multi-granularity prediction for scene text recognition}.
In: \bbtitle{Proceedings of the European Conference on Computer Vision},
pp. \bfpage{339}--\blpage{355}
(\byear{2022})
\end{bchapter}
\endbibitem

\bibitem[\protect\citeauthoryear{Croitoru et~al.}{2023}]{croitoru2023diffusion}
\begin{barticle}
\bauthor{\bsnm{Croitoru}, \binits{F.-A.}},
\bauthor{\bsnm{Hondru}, \binits{V.}},
\bauthor{\bsnm{Ionescu}, \binits{R.T.}},
\bauthor{\bsnm{Shah}, \binits{M.}}:
\batitle{Diffusion models in vision: A survey}.
\bjtitle{IEEE Transactions on Pattern Analysis and Machine Intelligence}
\bvolume{45}(\bissue{9}),
\bfpage{10850}--\blpage{10869}
(\byear{2023})
\end{barticle}
\endbibitem

\bibitem[\protect\citeauthoryear{Yang et~al.}{2022}]{yang2022diffusion}
\begin{botherref}
\oauthor{\bsnm{Yang}, \binits{L.}},
\oauthor{\bsnm{Zhang}, \binits{Z.}},
\oauthor{\bsnm{Song}, \binits{Y.}},
\oauthor{\bsnm{Hong}, \binits{S.}},
\oauthor{\bsnm{Xu}, \binits{R.}},
\oauthor{\bsnm{Zhao}, \binits{Y.}},
\oauthor{\bsnm{Shao}, \binits{Y.}},
\oauthor{\bsnm{Zhang}, \binits{W.}},
\oauthor{\bsnm{Cui}, \binits{B.}},
\oauthor{\bsnm{Yang}, \binits{M.-H.}}:
Diffusion models: A comprehensive survey of methods and applications.
arXiv preprint arXiv:2209.00796
(2022)
\end{botherref}
\endbibitem

\bibitem[\protect\citeauthoryear{Ho et~al.}{2020}]{ho2020denoising}
\begin{barticle}
\bauthor{\bsnm{Ho}, \binits{J.}},
\bauthor{\bsnm{Jain}, \binits{A.}},
\bauthor{\bsnm{Abbeel}, \binits{P.}}:
\batitle{Denoising diffusion probabilistic models}.
\bjtitle{Advances in Neural Information Processing Systems}
\bvolume{33},
\bfpage{6840}--\blpage{6851}
(\byear{2020})
\end{barticle}
\endbibitem

\bibitem[\protect\citeauthoryear{Song et~al.}{2021}]{song2020denoising}
\begin{bchapter}
\bauthor{\bsnm{Song}, \binits{J.}},
\bauthor{\bsnm{Meng}, \binits{C.}},
\bauthor{\bsnm{Ermon}, \binits{S.}}:
\bctitle{Denoising diffusion implicit models}.
In: \bbtitle{International Conference on Learning Representations},
pp. \bfpage{1}--\blpage{20}
(\byear{2021})
\end{bchapter}
\endbibitem

\bibitem[\protect\citeauthoryear{Zhang et~al.}{2023}]{zhang2022gddim}
\begin{bchapter}
\bauthor{\bsnm{Zhang}, \binits{Q.}},
\bauthor{\bsnm{Tao}, \binits{M.}},
\bauthor{\bsnm{Chen}, \binits{Y.}}:
\bctitle{g{DDIM}: Generalized denoising diffusion implicit models}.
In: \bbtitle{International Conference on Learning Representations},
pp. \bfpage{1}--\blpage{31}
(\byear{2023})
\end{bchapter}
\endbibitem

\bibitem[\protect\citeauthoryear{Nichol et~al.}{2022}]{nichol2021glide}
\begin{bchapter}
\bauthor{\bsnm{Nichol}, \binits{A.Q.}},
\bauthor{\bsnm{Dhariwal}, \binits{P.}},
\bauthor{\bsnm{Ramesh}, \binits{A.}},
\bauthor{\bsnm{Shyam}, \binits{P.}},
\bauthor{\bsnm{Mishkin}, \binits{P.}},
\bauthor{\bsnm{McGrew}, \binits{B.}},
\bauthor{\bsnm{Sutskever}, \binits{I.}},
\bauthor{\bsnm{Chen}, \binits{M.}}:
\bctitle{{GLIDE:} towards photorealistic image generation and editing with text-guided diffusion models}.
In: \bbtitle{International Conference on Machine Learning},
vol. \bseriesno{162},
pp. \bfpage{16784}--\blpage{16804}
(\byear{2022})
\end{bchapter}
\endbibitem

\bibitem[\protect\citeauthoryear{Ramesh et~al.}{2022}]{ramesh2022hierarchical}
\begin{botherref}
\oauthor{\bsnm{Ramesh}, \binits{A.}},
\oauthor{\bsnm{Dhariwal}, \binits{P.}},
\oauthor{\bsnm{Nichol}, \binits{A.}},
\oauthor{\bsnm{Chu}, \binits{C.}},
\oauthor{\bsnm{Chen}, \binits{M.}}:
Hierarchical text-conditional image generation with clip latents.
arXiv preprint arXiv:2204.06125
(2022)
\end{botherref}
\endbibitem

\bibitem[\protect\citeauthoryear{Saharia et~al.}{2022}]{saharia2022photorealistic}
\begin{barticle}
\bauthor{\bsnm{Saharia}, \binits{C.}},
\bauthor{\bsnm{Chan}, \binits{W.}},
\bauthor{\bsnm{Saxena}, \binits{S.}},
\bauthor{\bsnm{Li}, \binits{L.}},
\bauthor{\bsnm{Whang}, \binits{J.}},
\bauthor{\bsnm{Denton}, \binits{E.L.}},
\bauthor{\bsnm{Ghasemipour}, \binits{K.}},
\bauthor{\bsnm{Gontijo~Lopes}, \binits{R.}},
\bauthor{\bsnm{Karagol~Ayan}, \binits{B.}},
\bauthor{\bsnm{Salimans}, \binits{T.}}, \betal:
\batitle{Photorealistic text-to-image diffusion models with deep language understanding}.
\bjtitle{Advances in Neural Information Processing Systems}
\bvolume{35},
\bfpage{36479}--\blpage{36494}
(\byear{2022})
\end{barticle}
\endbibitem

\bibitem[\protect\citeauthoryear{Ho et~al.}{2022a}]{ho2022video}
\begin{botherref}
\oauthor{\bsnm{Ho}, \binits{J.}},
\oauthor{\bsnm{Salimans}, \binits{T.}},
\oauthor{\bsnm{Gritsenko}, \binits{A.}},
\oauthor{\bsnm{Chan}, \binits{W.}},
\oauthor{\bsnm{Norouzi}, \binits{M.}},
\oauthor{\bsnm{Fleet}, \binits{D.J.}}:
Video diffusion models.
arXiv preprint arXiv:2204.03458
(2022)
\end{botherref}
\endbibitem

\bibitem[\protect\citeauthoryear{Ho et~al.}{2022b}]{ho2022imagen}
\begin{botherref}
\oauthor{\bsnm{Ho}, \binits{J.}},
\oauthor{\bsnm{Chan}, \binits{W.}},
\oauthor{\bsnm{Saharia}, \binits{C.}},
\oauthor{\bsnm{Whang}, \binits{J.}},
\oauthor{\bsnm{Gao}, \binits{R.}},
\oauthor{\bsnm{Gritsenko}, \binits{A.}},
\oauthor{\bsnm{Kingma}, \binits{D.P.}},
\oauthor{\bsnm{Poole}, \binits{B.}},
\oauthor{\bsnm{Norouzi}, \binits{M.}},
\oauthor{\bsnm{Fleet}, \binits{D.J.}}, et al.:
Imagen video: High definition video generation with diffusion models.
arXiv preprint arXiv:2210.02303
(2022)
\end{botherref}
\endbibitem

\bibitem[\protect\citeauthoryear{Kong et~al.}{2021}]{kong2020diffwave}
\begin{bchapter}
\bauthor{\bsnm{Kong}, \binits{Z.}},
\bauthor{\bsnm{Ping}, \binits{W.}},
\bauthor{\bsnm{Huang}, \binits{J.}},
\bauthor{\bsnm{Zhao}, \binits{K.}},
\bauthor{\bsnm{Catanzaro}, \binits{B.}}:
\bctitle{Diffwave: A versatile diffusion model for audio synthesis}.
In: \bbtitle{International Conference on Learning Representations},
pp. \bfpage{1}--\blpage{17}
(\byear{2021})
\end{bchapter}
\endbibitem

\bibitem[\protect\citeauthoryear{Xu et~al.}{2023}]{xu2022dream3d}
\begin{bchapter}
\bauthor{\bsnm{Xu}, \binits{J.}},
\bauthor{\bsnm{Wang}, \binits{X.}},
\bauthor{\bsnm{Cheng}, \binits{W.}},
\bauthor{\bsnm{Cao}, \binits{Y.-P.}},
\bauthor{\bsnm{Shan}, \binits{Y.}},
\bauthor{\bsnm{Qie}, \binits{X.}},
\bauthor{\bsnm{Gao}, \binits{S.}}:
\bctitle{Dream3d: Zero-shot text-to-3d synthesis using 3d shape prior and text-to-image diffusion models}.
In: \bbtitle{Proceedings of the IEEE/CVF Conference on Computer Vision and Pattern Recognition},
pp. \bfpage{20908}--\blpage{20918}
(\byear{2023})
\end{bchapter}
\endbibitem

\bibitem[\protect\citeauthoryear{Li et~al.}{2022}]{li2022diffusion}
\begin{barticle}
\bauthor{\bsnm{Li}, \binits{X.}},
\bauthor{\bsnm{Thickstun}, \binits{J.}},
\bauthor{\bsnm{Gulrajani}, \binits{I.}},
\bauthor{\bsnm{Liang}, \binits{P.S.}},
\bauthor{\bsnm{Hashimoto}, \binits{T.B.}}:
\batitle{Diffusion-lm improves controllable text generation}.
\bjtitle{Advances in Neural Information Processing Systems}
\bvolume{35},
\bfpage{4328}--\blpage{4343}
(\byear{2022})
\end{barticle}
\endbibitem

\bibitem[\protect\citeauthoryear{Gong et~al.}{2023}]{gong2022diffuseq}
\begin{bchapter}
\bauthor{\bsnm{Gong}, \binits{S.}},
\bauthor{\bsnm{Li}, \binits{M.}},
\bauthor{\bsnm{Feng}, \binits{J.}},
\bauthor{\bsnm{Wu}, \binits{Z.}},
\bauthor{\bsnm{Kong}, \binits{L.}}:
\bctitle{{DiffuSeq}: Sequence to sequence text generation with diffusion models}.
In: \bbtitle{International Conference on Learning Representations},
pp. \bfpage{1}--\blpage{20}
(\byear{2023})
\end{bchapter}
\endbibitem

\bibitem[\protect\citeauthoryear{Sohl-Dickstein et~al.}{2015}]{sohl2015deep}
\begin{bchapter}
\bauthor{\bsnm{Sohl-Dickstein}, \binits{J.}},
\bauthor{\bsnm{Weiss}, \binits{E.}},
\bauthor{\bsnm{Maheswaranathan}, \binits{N.}},
\bauthor{\bsnm{Ganguli}, \binits{S.}}:
\bctitle{Deep unsupervised learning using nonequilibrium thermodynamics}.
In: \bbtitle{International Conference on Machine Learning},
pp. \bfpage{2256}--\blpage{2265}
(\byear{2015})
\end{bchapter}
\endbibitem

\bibitem[\protect\citeauthoryear{Esser et~al.}{2021}]{esser2021imagebart}
\begin{bchapter}
\bauthor{\bsnm{Esser}, \binits{P.}},
\bauthor{\bsnm{Rombach}, \binits{R.}},
\bauthor{\bsnm{Blattmann}, \binits{A.}},
\bauthor{\bsnm{Ommer}, \binits{B.}}:
\bctitle{{ImageBART: B}idirectional context with multinomial diffusion for autoregressive image synthesis}.
In: \beditor{\bsnm{Ranzato}, \binits{M.}},
\beditor{\bsnm{Beygelzimer}, \binits{A.}},
\beditor{\bsnm{Dauphin}, \binits{Y.}},
\beditor{\bsnm{Liang}, \binits{P.S.}},
\beditor{\bsnm{Vaughan}, \binits{J.W.}} (eds.)
\bbtitle{Advances in Neural Information Processing Systems},
vol. \bseriesno{34},
pp. \bfpage{3518}--\blpage{3532}
(\byear{2021})
\end{bchapter}
\endbibitem

\bibitem[\protect\citeauthoryear{Razavi et~al.}{2019}]{Razavi2019GeneratingDH}
\begin{bchapter}
\bauthor{\bsnm{Razavi}, \binits{A.}},
\bauthor{\bsnm{Oord}, \binits{A.}},
\bauthor{\bsnm{Vinyals}, \binits{O.}}:
\bctitle{Generating diverse high-fidelity images with {VQ-VAE-2}}.
In: \bbtitle{Neural Information Processing Systems},
pp. \bfpage{14837}--\blpage{14847}
(\byear{2019})
\end{bchapter}
\endbibitem

\bibitem[\protect\citeauthoryear{Gu et~al.}{2022}]{gu2022vector}
\begin{bchapter}
\bauthor{\bsnm{Gu}, \binits{S.}},
\bauthor{\bsnm{Chen}, \binits{D.}},
\bauthor{\bsnm{Bao}, \binits{J.}},
\bauthor{\bsnm{Wen}, \binits{F.}},
\bauthor{\bsnm{Zhang}, \binits{B.}},
\bauthor{\bsnm{Chen}, \binits{D.}},
\bauthor{\bsnm{Yuan}, \binits{L.}},
\bauthor{\bsnm{Guo}, \binits{B.}}:
\bctitle{Vector quantized diffusion model for text-to-image synthesis}.
In: \bbtitle{Proceedings of the IEEE/CVF Conference on Computer Vision and Pattern Recognition},
pp. \bfpage{10696}--\blpage{10706}
(\byear{2022})
\end{bchapter}
\endbibitem

\bibitem[\protect\citeauthoryear{Hoogeboom et~al.}{2021}]{hoogeboom2021argmax}
\begin{bchapter}
\bauthor{\bsnm{Hoogeboom}, \binits{E.}},
\bauthor{\bsnm{Nielsen}, \binits{D.}},
\bauthor{\bsnm{Jaini}, \binits{P.}},
\bauthor{\bsnm{Forr\'{e}}, \binits{P.}},
\bauthor{\bsnm{Welling}, \binits{M.}}:
\bctitle{Argmax flows and multinomial diffusion: Learning categorical distributions}.
In: \beditor{\bsnm{Ranzato}, \binits{M.}},
\beditor{\bsnm{Beygelzimer}, \binits{A.}},
\beditor{\bsnm{Dauphin}, \binits{Y.}},
\beditor{\bsnm{Liang}, \binits{P.S.}},
\beditor{\bsnm{Vaughan}, \binits{J.W.}} (eds.)
\bbtitle{Advances in Neural Information Processing Systems},
vol. \bseriesno{34},
pp. \bfpage{12454}--\blpage{12465}
(\byear{2021})
\end{bchapter}
\endbibitem

\bibitem[\protect\citeauthoryear{Austin et~al.}{2021}]{austin2021structured}
\begin{barticle}
\bauthor{\bsnm{Austin}, \binits{J.}},
\bauthor{\bsnm{Johnson}, \binits{D.D.}},
\bauthor{\bsnm{Ho}, \binits{J.}},
\bauthor{\bsnm{Tarlow}, \binits{D.}},
\bauthor{\bsnm{Van Den~Berg}, \binits{R.}}:
\batitle{Structured denoising diffusion models in discrete state-spaces}.
\bjtitle{Advances in Neural Information Processing Systems}
\bvolume{34},
\bfpage{17981}--\blpage{17993}
(\byear{2021})
\end{barticle}
\endbibitem

\bibitem[\protect\citeauthoryear{Zhu et~al.}{2022}]{zhu2022exploring}
\begin{botherref}
\oauthor{\bsnm{Zhu}, \binits{Z.}},
\oauthor{\bsnm{Wei}, \binits{Y.}},
\oauthor{\bsnm{Wang}, \binits{J.}},
\oauthor{\bsnm{Gan}, \binits{Z.}},
\oauthor{\bsnm{Zhang}, \binits{Z.}},
\oauthor{\bsnm{Wang}, \binits{L.}},
\oauthor{\bsnm{Hua}, \binits{G.}},
\oauthor{\bsnm{Wang}, \binits{L.}},
\oauthor{\bsnm{Liu}, \binits{Z.}},
\oauthor{\bsnm{Hu}, \binits{H.}}:
Exploring discrete diffusion models for image captioning.
arXiv preprint arXiv:2211.11694
(2022)
\end{botherref}
\endbibitem

\bibitem[\protect\citeauthoryear{He et~al.}{2016}]{he2016deep}
\begin{bchapter}
\bauthor{\bsnm{He}, \binits{K.}},
\bauthor{\bsnm{Zhang}, \binits{X.}},
\bauthor{\bsnm{Ren}, \binits{S.}},
\bauthor{\bsnm{Sun}, \binits{J.}}:
\bctitle{Deep residual learning for image recognition}.
In: \bbtitle{Proceedings of the IEEE/CVF Conference on Computer Vision and Pattern Recognition},
pp. \bfpage{770}--\blpage{778}
(\byear{2016})
\end{bchapter}
\endbibitem

\bibitem[\protect\citeauthoryear{Kenton and Toutanova}{2019}]{devlin2018bert}
\begin{bchapter}
\bauthor{\bsnm{Kenton}, \binits{J.D.M.-W.C.}},
\bauthor{\bsnm{Toutanova}, \binits{L.K.}}:
\bctitle{{BERT}: Pre-training of deep bidirectional transformers for language understanding}.
In: \bbtitle{Proceedings of Annual Conference of the North American Chapter of the Association for Computational Linguistics: Human Language Technologies},
pp. \bfpage{4171}--\blpage{4186}
(\byear{2019})
\end{bchapter}
\endbibitem

\bibitem[\protect\citeauthoryear{Ba et~al.}{2016}]{ba2016layer}
\begin{botherref}
\oauthor{\bsnm{Ba}, \binits{J.L.}},
\oauthor{\bsnm{Kiros}, \binits{J.R.}},
\oauthor{\bsnm{Hinton}, \binits{G.E.}}:
Layer normalization.
arXiv preprint arXiv:1607.06450
(2016)
\end{botherref}
\endbibitem

\bibitem[\protect\citeauthoryear{Gupta et~al.}{2016}]{gupta2016synthetic}
\begin{bchapter}
\bauthor{\bsnm{Gupta}, \binits{A.}},
\bauthor{\bsnm{Vedaldi}, \binits{A.}},
\bauthor{\bsnm{Zisserman}, \binits{A.}}:
\bctitle{Synthetic data for text localisation in natural images}.
In: \bbtitle{Proceedings of the IEEE/CVF Conference on Computer Vision and Pattern Recognition},
pp. \bfpage{2315}--\blpage{2324}
(\byear{2016})
\end{bchapter}
\endbibitem

\bibitem[\protect\citeauthoryear{Jiang et~al.}{2023}]{jiang2023revisiting}
\begin{bchapter}
\bauthor{\bsnm{Jiang}, \binits{Q.}},
\bauthor{\bsnm{Wang}, \binits{J.}},
\bauthor{\bsnm{Peng}, \binits{D.}},
\bauthor{\bsnm{Liu}, \binits{C.}},
\bauthor{\bsnm{Jin}, \binits{L.}}:
\bctitle{Revisiting scene text recognition: A data perspective}.
In: \bbtitle{Proceedings of the IEEE/CVF International Conference on Computer Vision},
pp. \bfpage{20543}--\blpage{20554}
(\byear{2023})
\end{bchapter}
\endbibitem

\bibitem[\protect\citeauthoryear{Sun et~al.}{2019}]{sun2019icdar}
\begin{bchapter}
\bauthor{\bsnm{Sun}, \binits{Y.}},
\bauthor{\bsnm{Ni}, \binits{Z.}},
\bauthor{\bsnm{Chng}, \binits{C.-K.}},
\bauthor{\bsnm{Liu}, \binits{Y.}},
\bauthor{\bsnm{Luo}, \binits{C.}},
\bauthor{\bsnm{Ng}, \binits{C.C.}},
\bauthor{\bsnm{Han}, \binits{J.}},
\bauthor{\bsnm{Ding}, \binits{E.}},
\bauthor{\bsnm{Liu}, \binits{J.}},
\bauthor{\bsnm{Karatzas}, \binits{D.}}, \betal:
\bctitle{{ICDAR} 2019 competition on large-scale street view text with partial labeling -- {RRC-LSVT}}.
In: \bbtitle{International Conference on Document Analysis and Recognition},
pp. \bfpage{1557}--\blpage{1562}
(\byear{2019})
\end{bchapter}
\endbibitem

\bibitem[\protect\citeauthoryear{Veit et~al.}{2016}]{veit2016coco}
\begin{botherref}
\oauthor{\bsnm{Veit}, \binits{A.}},
\oauthor{\bsnm{Matera}, \binits{T.}},
\oauthor{\bsnm{Neumann}, \binits{L.}},
\oauthor{\bsnm{Matas}, \binits{J.}},
\oauthor{\bsnm{Belongie}, \binits{S.}}:
{COCO-Text: Dataset} and benchmark for text detection and recognition in natural images.
arXiv preprint arXiv:1601.07140
(2016)
\end{botherref}
\endbibitem

\bibitem[\protect\citeauthoryear{Zhang et~al.}{2017}]{zhang2017uber}
\begin{bchapter}
\bauthor{\bsnm{Zhang}, \binits{Y.}},
\bauthor{\bsnm{Gueguen}, \binits{L.}},
\bauthor{\bsnm{Zharkov}, \binits{I.}},
\bauthor{\bsnm{Zhang}, \binits{P.}},
\bauthor{\bsnm{Seifert}, \binits{K.}},
\bauthor{\bsnm{Kadlec}, \binits{B.}}:
\bctitle{{Uber-Text: A} large-scale dataset for optical character recognition from street-level imagery}.
In: \bbtitle{SUNw: Scene Understanding Workshop-CVPR},
p. \bfpage{5}
(\byear{2017})
\end{bchapter}
\endbibitem

\bibitem[\protect\citeauthoryear{Xie et~al.}{2022}]{xie2022toward}
\begin{bchapter}
\bauthor{\bsnm{Xie}, \binits{X.}},
\bauthor{\bsnm{Fu}, \binits{L.}},
\bauthor{\bsnm{Zhang}, \binits{Z.}},
\bauthor{\bsnm{Wang}, \binits{Z.}},
\bauthor{\bsnm{Bai}, \binits{X.}}:
\bctitle{{Toward understanding WordArt: Corner-guided transformer for scene text recognition}}.
In: \bbtitle{Proceedings of the European Conference on Computer Vision},
pp. \bfpage{303}--\blpage{321}
(\byear{2022})
\end{bchapter}
\endbibitem

\bibitem[\protect\citeauthoryear{Karatzas et~al.}{2013}]{karatzas2013icdar}
\begin{bchapter}
\bauthor{\bsnm{Karatzas}, \binits{D.}},
\bauthor{\bsnm{Shafait}, \binits{F.}},
\bauthor{\bsnm{Uchida}, \binits{S.}},
\bauthor{\bsnm{Iwamura}, \binits{M.}},
\bauthor{\bsnm{Bigorda}, \binits{L.G.}},
\bauthor{\bsnm{Mestre}, \binits{S.R.}},
\bauthor{\bsnm{Mas}, \binits{J.}},
\bauthor{\bsnm{Mota}, \binits{D.F.}},
\bauthor{\bsnm{Almazan}, \binits{J.A.}},
\bauthor{\bsnm{De~Las~Heras}, \binits{L.P.}}:
\bctitle{{ICDAR 2013 robust reading competition}}.
In: \bbtitle{International Conference on Document Analysis and Recognition},
pp. \bfpage{1484}--\blpage{1493}
(\byear{2013})
\end{bchapter}
\endbibitem

\bibitem[\protect\citeauthoryear{Karatzas et~al.}{2015}]{karatzas2015icdar}
\begin{bchapter}
\bauthor{\bsnm{Karatzas}, \binits{D.}},
\bauthor{\bsnm{Gomez-Bigorda}, \binits{L.}},
\bauthor{\bsnm{Nicolaou}, \binits{A.}},
\bauthor{\bsnm{Ghosh}, \binits{S.}},
\bauthor{\bsnm{Bagdanov}, \binits{A.}},
\bauthor{\bsnm{Iwamura}, \binits{M.}},
\bauthor{\bsnm{Matas}, \binits{J.}},
\bauthor{\bsnm{Neumann}, \binits{L.}},
\bauthor{\bsnm{Chandrasekhar}, \binits{V.R.}},
\bauthor{\bsnm{Lu}, \binits{S.}}, \betal:
\bctitle{{ICDAR} 2015 competition on robust reading}.
In: \bbtitle{International Conference on Document Analysis and Recognition},
pp. \bfpage{1156}--\blpage{1160}
(\byear{2015})
\end{bchapter}
\endbibitem

\bibitem[\protect\citeauthoryear{Phan et~al.}{2013}]{phan2013recognizing}
\begin{bchapter}
\bauthor{\bsnm{Phan}, \binits{T.Q.}},
\bauthor{\bsnm{Shivakumara}, \binits{P.}},
\bauthor{\bsnm{Tian}, \binits{S.}},
\bauthor{\bsnm{Tan}, \binits{C.L.}}:
\bctitle{Recognizing text with perspective distortion in natural scenes}.
In: \bbtitle{Proceedings of the IEEE International Conference on Computer Vision},
pp. \bfpage{569}--\blpage{576}
(\byear{2013})
\end{bchapter}
\endbibitem

\bibitem[\protect\citeauthoryear{Risnumawan et~al.}{2014}]{risnumawan2014robust}
\begin{barticle}
\bauthor{\bsnm{Risnumawan}, \binits{A.}},
\bauthor{\bsnm{Shivakumara}, \binits{P.}},
\bauthor{\bsnm{Chan}, \binits{C.S.}},
\bauthor{\bsnm{Tan}, \binits{C.L.}}:
\batitle{A robust arbitrary text detection system for natural scene images}.
\bjtitle{Expert Systems with Applications}
\bvolume{41}(\bissue{18}),
\bfpage{8027}--\blpage{8048}
(\byear{2014})
\end{barticle}
\endbibitem

\bibitem[\protect\citeauthoryear{Chen et~al.}{2021}]{chen2021benchmarking}
\begin{botherref}
\oauthor{\bsnm{Chen}, \binits{J.}},
\oauthor{\bsnm{Yu}, \binits{H.}},
\oauthor{\bsnm{Ma}, \binits{J.}},
\oauthor{\bsnm{Guan}, \binits{M.}},
\oauthor{\bsnm{Xu}, \binits{X.}},
\oauthor{\bsnm{Wang}, \binits{X.}},
\oauthor{\bsnm{Qu}, \binits{S.}},
\oauthor{\bsnm{Li}, \binits{B.}},
\oauthor{\bsnm{Xue}, \binits{X.}}:
Benchmarking chinese text recognition: Datasets, baselines, and an empirical study.
arXiv preprint arXiv:2112.15093
(2021)
\end{botherref}
\endbibitem

\bibitem[\protect\citeauthoryear{Shi et~al.}{2017}]{shi2017icdar2017}
\begin{bchapter}
\bauthor{\bsnm{Shi}, \binits{B.}},
\bauthor{\bsnm{Yao}, \binits{C.}},
\bauthor{\bsnm{Liao}, \binits{M.}},
\bauthor{\bsnm{Yang}, \binits{M.}},
\bauthor{\bsnm{Xu}, \binits{P.}},
\bauthor{\bsnm{Cui}, \binits{L.}},
\bauthor{\bsnm{Belongie}, \binits{S.}},
\bauthor{\bsnm{Lu}, \binits{S.}},
\bauthor{\bsnm{Bai}, \binits{X.}}:
\bctitle{{ICDAR2017 competition on reading chinese text in the wild (RCTW-17)}}.
In: \bbtitle{International Conference on Document Analysis and Recognition},
vol. \bseriesno{1},
pp. \bfpage{1429}--\blpage{1434}
(\byear{2017})
\end{bchapter}
\endbibitem

\bibitem[\protect\citeauthoryear{Zhang et~al.}{2019}]{zhang2019icdar}
\begin{bchapter}
\bauthor{\bsnm{Zhang}, \binits{R.}},
\bauthor{\bsnm{Zhou}, \binits{Y.}},
\bauthor{\bsnm{Jiang}, \binits{Q.}},
\bauthor{\bsnm{Song}, \binits{Q.}},
\bauthor{\bsnm{Li}, \binits{N.}},
\bauthor{\bsnm{Zhou}, \binits{K.}},
\bauthor{\bsnm{Wang}, \binits{L.}},
\bauthor{\bsnm{Wang}, \binits{D.}},
\bauthor{\bsnm{Liao}, \binits{M.}},
\bauthor{\bsnm{Yang}, \binits{M.}}, \betal:
\bctitle{{ICDAR} 2019 robust reading challenge on reading chinese text on signboard}.
In: \bbtitle{International Conference on Document Analysis and Recognition},
pp. \bfpage{1577}--\blpage{1581}
(\byear{2019})
\end{bchapter}
\endbibitem

\bibitem[\protect\citeauthoryear{Chng et~al.}{2019}]{chng2019icdar2019}
\begin{bchapter}
\bauthor{\bsnm{Chng}, \binits{C.K.}},
\bauthor{\bsnm{Liu}, \binits{Y.}},
\bauthor{\bsnm{Sun}, \binits{Y.}},
\bauthor{\bsnm{Ng}, \binits{C.C.}},
\bauthor{\bsnm{Luo}, \binits{C.}},
\bauthor{\bsnm{Ni}, \binits{Z.}},
\bauthor{\bsnm{Fang}, \binits{C.}},
\bauthor{\bsnm{Zhang}, \binits{S.}},
\bauthor{\bsnm{Han}, \binits{J.}},
\bauthor{\bsnm{Ding}, \binits{E.}}, \betal:
\bctitle{{ICDAR}2019 robust reading challenge on arbitrary-shaped text {(RRC-ArT)}}.
In: \bbtitle{International Conference on Document Analysis and Recognition},
pp. \bfpage{1571}--\blpage{1576}
(\byear{2019})
\end{bchapter}
\endbibitem

\bibitem[\protect\citeauthoryear{Yuan et~al.}{2019}]{yuan2019large}
\begin{barticle}
\bauthor{\bsnm{Yuan}, \binits{T.-L.}},
\bauthor{\bsnm{Zhu}, \binits{Z.}},
\bauthor{\bsnm{Xu}, \binits{K.}},
\bauthor{\bsnm{Li}, \binits{C.-J.}},
\bauthor{\bsnm{Mu}, \binits{T.-J.}},
\bauthor{\bsnm{Hu}, \binits{S.-M.}}:
\batitle{A large chinese text dataset in the wild}.
\bjtitle{Journal of Computer Science and Technology}
\bvolume{34},
\bfpage{509}--\blpage{521}
(\byear{2019})
\end{barticle}
\endbibitem

\bibitem[\protect\citeauthoryear{He et~al.}{2018}]{he2018icpr2018}
\begin{bchapter}
\bauthor{\bsnm{He}, \binits{M.}},
\bauthor{\bsnm{Liu}, \binits{Y.}},
\bauthor{\bsnm{Yang}, \binits{Z.}},
\bauthor{\bsnm{Zhang}, \binits{S.}},
\bauthor{\bsnm{Luo}, \binits{C.}},
\bauthor{\bsnm{Gao}, \binits{F.}},
\bauthor{\bsnm{Zheng}, \binits{Q.}},
\bauthor{\bsnm{Wang}, \binits{Y.}},
\bauthor{\bsnm{Zhang}, \binits{X.}},
\bauthor{\bsnm{Jin}, \binits{L.}}:
\bctitle{{ICPR2018 contest on robust reading for multi-type web images}}.
In: \bbtitle{International Conference on Pattern Recognition},
pp. \bfpage{7}--\blpage{12}
(\byear{2018})
\end{bchapter}
\endbibitem

\bibitem[\protect\citeauthoryear{Zhang et~al.}{2020}]{zhang2020scut}
\begin{barticle}
\bauthor{\bsnm{Zhang}, \binits{H.}},
\bauthor{\bsnm{Liang}, \binits{L.}},
\bauthor{\bsnm{Jin}, \binits{L.}}:
\batitle{{SCUT-HCCDoc: A} new benchmark dataset of handwritten chinese text in unconstrained camera-captured documents}.
\bjtitle{Pattern Recognition}
\bvolume{108},
\bfpage{107559}
(\byear{2020})
\end{barticle}
\endbibitem

\bibitem[\protect\citeauthoryear{Kingma and Ba}{2015}]{kingma2015adam}
\begin{bchapter}
\bauthor{\bsnm{Kingma}, \binits{D.P.}},
\bauthor{\bsnm{Ba}, \binits{J.}}:
\bctitle{Adam: {A} method for stochastic optimization}.
In: \bbtitle{International Conference on Learning Representations},
pp. \bfpage{1}--\blpage{15}
(\byear{2015})
\end{bchapter}
\endbibitem

\bibitem[\protect\citeauthoryear{Smith and Topin}{2019}]{smith2019super}
\begin{bchapter}
\bauthor{\bsnm{Smith}, \binits{L.N.}},
\bauthor{\bsnm{Topin}, \binits{N.}}:
\bctitle{Super-convergence: Very fast training of neural networks using large learning rates}.
In: \bbtitle{Artificial Intelligence and Machine Learning for Multi-Domain Operations Applications},
vol. \bseriesno{11006},
pp. \bfpage{369}--\blpage{386}
(\byear{2019})
\end{bchapter}
\endbibitem

\bibitem[\protect\citeauthoryear{Izmailov et~al.}{2018}]{izmailov2018avergaingwl}
\begin{bchapter}
\bauthor{\bsnm{Izmailov}, \binits{P.}},
\bauthor{\bsnm{Podoprikhin}, \binits{D.}},
\bauthor{\bsnm{Garipov}, \binits{T.}},
\bauthor{\bsnm{Vetrov}, \binits{D.P.}},
\bauthor{\bsnm{Wilson}, \binits{A.G.}}:
\bctitle{Averaging weights leads to wider optima and better generalization}.
In: \bbtitle{Conference on Uncertainty in Artificial Intelligence}
(\byear{2018})
\end{bchapter}
\endbibitem

\bibitem[\protect\citeauthoryear{Na et~al.}{2022}]{na2022multi}
\begin{bchapter}
\bauthor{\bsnm{Na}, \binits{B.}},
\bauthor{\bsnm{Kim}, \binits{Y.}},
\bauthor{\bsnm{Park}, \binits{S.}}:
\bctitle{Multi-modal text recognition networks: Interactive enhancements between visual and semantic features}.
In: \bbtitle{Proceedings of the European Conference on Computer Vision},
pp. \bfpage{446}--\blpage{463}
(\byear{2022})
\end{bchapter}
\endbibitem

\bibitem[\protect\citeauthoryear{Merity et~al.}{2017}]{merity2017pointer}
\begin{bchapter}
\bauthor{\bsnm{Merity}, \binits{S.}},
\bauthor{\bsnm{Xiong}, \binits{C.}},
\bauthor{\bsnm{Bradbury}, \binits{J.}},
\bauthor{\bsnm{Socher}, \binits{R.}}:
\bctitle{Pointer sentinel mixture models}.
In: \bbtitle{International Conference on Learning Representations}
(\byear{2017})
\end{bchapter}
\endbibitem

\bibitem[\protect\citeauthoryear{Yue et~al.}{2020}]{yue2020robustscanner}
\begin{bchapter}
\bauthor{\bsnm{Yue}, \binits{X.}},
\bauthor{\bsnm{Kuang}, \binits{Z.}},
\bauthor{\bsnm{Lin}, \binits{C.}},
\bauthor{\bsnm{Sun}, \binits{H.}},
\bauthor{\bsnm{Zhang}, \binits{W.}}:
\bctitle{{RobustScanner: D}ynamically enhancing positional clues for robust text recognition}.
In: \bbtitle{Proceedings of the European Conference on Computer Vision},
pp. \bfpage{135}--\blpage{151}
(\byear{2020})
\end{bchapter}
\endbibitem

\bibitem[\protect\citeauthoryear{Lu et~al.}{2021}]{lu2021master}
\begin{barticle}
\bauthor{\bsnm{Lu}, \binits{N.}},
\bauthor{\bsnm{Yu}, \binits{W.}},
\bauthor{\bsnm{Qi}, \binits{X.}},
\bauthor{\bsnm{Chen}, \binits{Y.}},
\bauthor{\bsnm{Gong}, \binits{P.}},
\bauthor{\bsnm{Xiao}, \binits{R.}},
\bauthor{\bsnm{Bai}, \binits{X.}}:
\batitle{{MASTER: Multi-aspect non-local network for scene text recognition}}.
\bjtitle{Pattern Recognition}
\bvolume{117},
\bfpage{107980}
(\byear{2021})
\end{barticle}
\endbibitem

\bibitem[\protect\citeauthoryear{Chen et~al.}{2021}]{chen2021scene}
\begin{bchapter}
\bauthor{\bsnm{Chen}, \binits{J.}},
\bauthor{\bsnm{Li}, \binits{B.}},
\bauthor{\bsnm{Xue}, \binits{X.}}:
\bctitle{Scene text telescope: Text-focused scene image super-resolution}.
In: \bbtitle{Proceedings of the IEEE/CVF Conference on Computer Vision and Pattern Recognition},
pp. \bfpage{12026}--\blpage{12035}
(\byear{2021})
\end{bchapter}
\endbibitem

\bibitem[\protect\citeauthoryear{Lyu et~al.}{2022}]{lyu2022maskocr}
\begin{botherref}
\oauthor{\bsnm{Lyu}, \binits{P.}},
\oauthor{\bsnm{Zhang}, \binits{C.}},
\oauthor{\bsnm{Liu}, \binits{S.}},
\oauthor{\bsnm{Qiao}, \binits{M.}},
\oauthor{\bsnm{Xu}, \binits{Y.}},
\oauthor{\bsnm{Wu}, \binits{L.}},
\oauthor{\bsnm{Yao}, \binits{K.}},
\oauthor{\bsnm{Han}, \binits{J.}},
\oauthor{\bsnm{Ding}, \binits{E.}},
\oauthor{\bsnm{Wang}, \binits{J.}}:
{MaskOCR: T}ext recognition with masked encoder-decoder pretraining.
arXiv preprint arXiv:2206.00311
(2022)
\end{botherref}
\endbibitem

\end{thebibliography}

\end{document}